\documentclass[pdflatex,sn-mathphys-num]{sn-jnl}

\usepackage{graphicx}%
\usepackage{multirow}%
\usepackage{amsmath,amssymb,amsfonts}%
\usepackage{amsthm}%
\usepackage{mathrsfs}%
\usepackage[title]{appendix}%
\usepackage{xcolor}%
\usepackage{textcomp}%
\usepackage{manyfoot}%
\usepackage{booktabs}%
\usepackage{algorithm}%
\usepackage{algorithmicx}%
\usepackage{algpseudocode}%
\usepackage{listings}%
\usepackage{graphicx}
\usepackage{float}
\usepackage{arydshln}
\usepackage{pifont}
\usepackage{subcaption}
\usepackage{threeparttable}
\usepackage{multirow}
\newenvironment{revision}
{}
{}
\newcommand{\revisiontext}[1]{#1}

\theoremstyle{thmstyleone}%
\theoremstyle{thmstyletwo}%

\theoremstyle{thmstylethree}%

\begin{document}

\title[Article Title]{\revisiontext{Mitigating Task Objective Conflicts for Continual Learning in Unified Multimodal Models}}


\author[1]{\fnm{Jiaxing} \sur{Zhang}}\email{zjxing9972@gmail.com}

\author*[2]{\fnm{Hao} \sur{Tang}}\email{haotang@pku.edu.cn}


\affil[1]{\orgdiv{School of Computer Science}, 
\orgname{Shanghai Jiao Tong University}, 
\orgaddress{\street{No. 800 Dongchuan Road}, 
\city{Shanghai}, 
\postcode{200240}, 
\state{Shanghai}, 
\country{China}}}

\affil*[2]{\orgdiv{School of Computer Science}, 
\orgname{Peking University}, 
\orgaddress{
\street{No. 5 Yiheyuan Road}, 
\city{Beijing}, 
\postcode{100871}, 
\state{Beijing}, 
\country{China}
}}



\abstract{\revisiontext{Recent advances in unified autoregressive (AR) models have enabled multimodal understanding (MMU) and text-to-image generation (T2I) within a single architecture. However, in continual learning settings, their divergent objectives can induce task interference, hindering knowledge preservation and adaptation to new tasks. In this work, we investigate \textbf{T}ask \textbf{O}bjective \textbf{C}onflict (TOC) through theoretical and empirical analyses, revealing conflicting optimization directions at the task level but complementary interactions at the token level. These findings suggest that continual adaptation requires mitigating task interference while preserving beneficial cross-task knowledge sharing. To this end, we propose \textbf{UniDecouple}, a selective decoupling framework based on a \textbf{T}ask-\textbf{A}ware \textbf{M}ixture-\textbf{o}f-\textbf{E}xperts (TA-MoE) architecture. Specifically, UniDecouple retains shared attention layers to preserve common contextual dependencies while introducing task-specific feed-forward network (FFN) experts to mitigate optimization conflicts. Furthermore, hierarchical expert routing enables adaptive expert selection, while hybrid expert collaboration leverages shared experts to facilitate cross-task knowledge transfer. A two-stage training strategy further balances task specialization and coordinated adaptation. Extensive experiments and ablation studies demonstrate that UniDecouple improves both multimodal understanding and image generation performance under a comparable per-token activation budget, achieving a favorable balance between knowledge preservation and task adaptation in continual learning.}}




\maketitle

\section{Introduction}\label{sec1}

Human cognition, where understanding and generation intricately interleave, provides the foundation for reasoning and thought~\cite{islam2025gpt,wu2024janus,tong2024metamorph}. Inspired by this mechanism, researchers have sought unified models that integrate multimodal understanding and generation within a single framework. Yet progress remains divided: autoregressive (AR) based models~\cite{zhu2024llava,liu2024llavanext,zhu2023minigpt,bai2023qwen} drive multimodal understanding (MMU), while diffusion approaches~\cite{podell2023sdxl,wang2024emu3,rombach2022high} dominate text-to-image generation (T2I). To bridge these divergent paradigms, recent works (e.g., DALL·E~\cite{reddy2021dall}) employ AR based models for T2I, resulting in a preliminarily unified paradigm and prompting further investigations~\cite{team2024chameleon,chern2024anole,dong2023dreamllm,wang2024illume,wu2024vila} into extending AR frameworks to support both MMU and T2I. Yet, \revisiontext{continually adapting unified AR models to MMU and T2I tasks remains challenging, as divergent task objectives can induce interference and compromise previously acquired capabilities.}

A widely recognized challenge for AR based unified models arises from the representation level~\cite{wu2024janus,ma2024janusflow,chen2025janus}. MMU relies on semantic abstraction to capture high-level information, whereas T2I demands fine-grained detail preservation to faithfully render the image. The conflict in representation requirements has been widely recognized, and to reconcile it, early approaches first adopted a single pixel or semantic encoder to learn shared representations~\cite{wu2024vila,lin2025toklip}. Nevertheless, relying on a single encoder often biases the extracted features toward one task and cannot adequately satisfy these distinct objectives. Subsequently, JanusPro~\cite{chen2025janus} decouples the encoders, leveraging both pixel and semantic encoders to derive task-specific representations, effectively resolving the representation conflict.

Nevertheless, prior work on unifying MMU and T2I within AR based architectures focuses on surface-level representation conflicts, overlooking the deeper issues inherent in the architecture itself. Consequently, we focus on the internal information flow \revisiontext{and investigate potential optimization conflicts arising from shared computational dependencies}~\cite{sener2018multi,yu2020gradient}. Subsequently, we conduct theoretical \revisiontext{analyses} and empirical experiments to validate this hypothesis, demonstrating that these conflicts, termed Task Objective Conflict (TOC), indeed exist. Specifically, \revisiontext{joint optimization of MMU and T2I can induce negative transfer, where updates benefiting one task adversely affect the other. Our gradient-based analysis further reveals that, although the two tasks exhibit conflicting optimization directions at the task-level, complementary interactions coexist at token-level. This suggests that effective continual adaptation requires mitigating task interference while preserving beneficial cross-task knowledge sharing.}

\begin{figure*}
    \centering
    \includegraphics[width=1\linewidth,trim={14.3cm 6.2cm 9cm 5.2cm},clip]{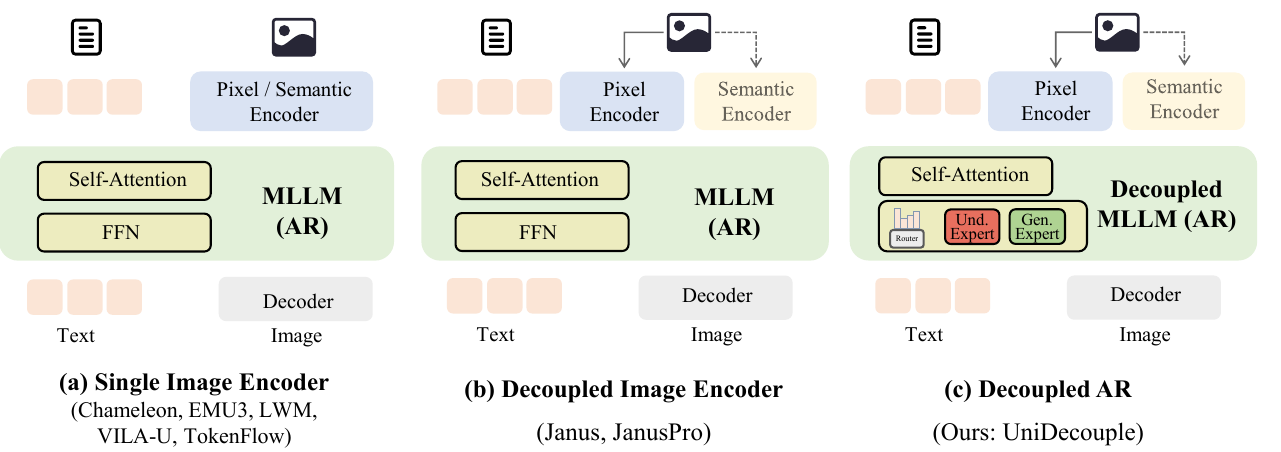}
    \caption{Comparison of different multimodal large language model (MLLM) architectures. (a) Single Image Encoder: pixel and semantic features are jointly encoded into a unified representation (e.g., Chameleon, EMU3, LWM, VILA-U, TokenFlow). (b) Decoupled Image Encoder: pixel-level and semantic-level features are separately processed, then integrated into the MLLM (e.g., Janus, JanusPro). (c) Decoupled AR (Ours): UniDecouple introduces task-specific experts via a Mixture-of-Experts (MoE) to resolve conflicts between \revisiontext{semantic-level understanding and pixel-level generation}.}
    \label{fig:intro}
\end{figure*}

To mitigate TOC, we propose UniDecouple. As illustrated in Fig.~\ref{fig:intro}(c), UniDecouple \revisiontext{adapts pretrained unified AR models through selective FFN-level decoupling based on a Mixture of Experts (MoE), while retaining shared attention layers to preserve common contextual dependencies and transferable knowledge. This design provides task-specific optimization capacity without completely separating cross-task interactions, making it suitable for continual learning.} Nevertheless, the conventional MoE architecture remains inadequate for \revisiontext{effectively mitigating} these conflicts, as it lacks explicit mechanisms to orchestrate task-specific experts and reconcile the divergent objectives of MMU and T2I. To this end, we design Task-Aware Mixture of Experts (TA-MoE), which incorporates key components: Hierarchical Expert Routing, enabling \revisiontext{adaptive token-level expert selection beyond fixed task assignments}, and Hybrid Expert Collaboration, facilitating the integration of complementary knowledge \revisiontext{through shared experts}. We also introduce a two-stage training strategy: training experts for task-specific skills and jointly fine-tuning to balance specialization and overall coordination.

We empirically validate the effectiveness of UniDecouple through a series of experiments. In Sec.~\ref{confirm}, we first validate our hypothesis by \revisiontext{directly measuring task-level gradient interference, analyzing token-level interactions, and examining the influence of causal masking on cross-task optimization}. In Sec.~\ref{compare}, we then conduct both quantitative and qualitative evaluations on mainstream understanding and generation benchmarks, demonstrating that UniDecouple consistently achieves strong performance across tasks. Finally, in Sec.~\ref{ablation}, ablation studies \revisiontext{evaluate the contributions of selective FFN decoupling, learnable routing, shared experts, and the two-stage training strategy, while examining model capacity and activated parameters}. Overall, these results demonstrate \revisiontext{that UniDecouple effectively balances task specialization, cross-task knowledge sharing, and knowledge preservation during continual adaptation}.

Our main contributions are summarized below:
\begin{itemize}
    \item We theoretically and empirically validate Task Objective Conflict (TOC) in AR models, \revisiontext{revealing task-level gradient interference and token-level complementarity.}

    \item We propose UniDecouple, a Task-Aware Mixture-of-Experts (TA-MoE) framework \revisiontext{that balances task specialization and cross-task knowledge sharing through selective FFN decoupling and adaptive expert routing in continual learning.}

    \item Our comprehensive experiments show that \revisiontext{UniDecouple improves both MMU and T2I performance with comparable per-token activated parameters, while effectively balancing knowledge preservation and task adaptation.}
\end{itemize}

\revisiontext{Part of this work appeared as an Extended Abstract at AAMAS 2026\footnote{Paper link: \url{https://www.ifaamas.org/Proceedings/aamas2026/pdfs/DLJD3800.pdf}} \cite{zhang2025resolving}. Compared with the brief conference presentation, this journal version expands the theoretical analysis of TOC and gradient interference in Sec.~\ref{subsec:conflict}, and provides detailed formulations of hierarchical routing, shared-expert collaboration, and two-stage training in Secs.~\ref{tamoe} and~\ref{train}, including sparse expert activation and LoRA-based parameter-efficient optimization. The experimental evaluation is broadened with additional benchmarks, including DPG-Bench, and comparisons with understanding-only, generation-only, and unified models in Sec.~\ref{compare} (Tables~\ref{compare-understanding}--\ref{dpg}). We further analyze expert utilization through activation/load distributions in Sec.~\ref{expert} (Fig.~\ref{fig:expert-load}). Finally, Sec.~\ref{ablation} (Tables~\ref{tab:albation1}--\ref{tab:ratio}) presents detailed ablation results for the Task-Aware Router with group supervision, Shared Expert, training strategy, and expert ratio, with Table~\ref{tab:albation2} reporting training epochs and task performance to assess training convergence.}


\section{Related Work}

\noindent\textbf{Mixture of Experts (MoE).}
Recent studies~\cite{dai2024deepseekmoe,liu2024deepseek,cai2024survey,yang2025fairsmoe} have demonstrated the effectiveness of MoE when integrated with large language models. For instance, Mixtral-MoE 8x7B~\cite{jiang2024mixtral} excels in scaling model capacity and improving task performance. DeepSeek introduced Top-K routing and shared experts, enhancing MoE’s task allocation efficiency. Variants such as Chen et al.~\cite{chen2023adamv} enable multitask learning for simple tasks through embedding analysis, while Li et al.~\cite{li2025uni} assign different experts for different modalities, proving the feasibility of expert specialization.

\noindent\textbf{Multimodal Understanding (MMU).} 
Multimodal large language models (MLLMs), such as Flamingo~\cite{alayrac2022flamingo}, LLaVA~\cite{liu2023visual}, MiniGPT-4~\cite{zhu2023minigpt}, BLIP-2~\cite{li2023blip}, InstructBLIP~\cite{panagopoulou2023x}, Qwen2-VL~\cite{wang2024qwen2}, and Intern-VL~\cite{chen2024internvl}, demonstrate strong capabilities in processing multimodal information. They typically align features from pretrained modality-specific encoders (e.g., CLIP~\cite{radford2021learning}) using methods such as Q-Former and connectors, and then map them into the LLM to enable cross-modal reasoning. Recent studies~\cite{ge2023planting,jin2023unified,sun2023emu} have attempted to extend MLLMs beyond perception to generation and other domains, opening new research directions.

\noindent\textbf{Text-to-Image Generation (T2I).} 
T2I generation has undergone a series of methodological advancements \cite{tao2022df,peebles2023scalable,tao2023galip,lipman2022flow,reddy2021dall}. In particular, diffusion-based methods generate high-quality samples by iteratively denoising Gaussian perturbations in continuous latent spaces~\cite{ho2020denoising,nichol2021improved}. The traditional denoising process relied on U-Net; it has gradually evolved to a Transformer-based prediction method~\cite{peebles2023scalable}. Diffusion models excel in generation fidelity and stability, but their multi-step iterative process results in slow inference. To improve efficiency, flow matching~\cite{lipman2022flow} generative models were introduced, learning continuous probability flows to more directly approximate the data distribution while retaining the advantages of diffusion. Subsequently, autoregressive (AR) models began to tackle visual generation tasks~\cite{reddy2021dall}, discretizing images into token sequences and employing Transformers to predict each token sequentially. These advancements lay a solid foundation for the development of subsequent unified multimodal models.

\noindent\textbf{Unified Multimodal Models.}
Unified multimodal models based on autoregressive (AR) architectures enable multimodal understanding (MMU) and T2I generation within a single framework~\cite{team2024chameleon,zhou2024transfusion}. However, existing approaches, whether employing shared encoders~\cite{wu2024vila,team2024chameleon} or decoupled ones~\cite{wu2024janus,ma2024janusflow,chen2025janus}, remain susceptible to task objective conflicts (TOC), leading to negative transfer during joint optimization.
\revisiontext{To improve task specialization, VLMo~\cite{bao2022vlmo} and Uni-MoE~\cite{li2025uni} introduce modality-specific FFN experts with shared attention, while Mixture-of-Transformers (MoT)~\cite{liang2024mixture} further decouples attention projections and normalization layers. Recent models, including BAGEL~\cite{deng2025emerging} and Mogao~\cite{liao2025mogao}, employ modality-specific parameters for unified MMU and T2I, while LMFusion~\cite{shi2026lmfusion} explores preserving pretrained language capabilities when incorporating visual generation. Nevertheless, these methods primarily focus on multimodal pretraining or capability expansion, rather than continual adaptation of models already supporting both tasks. In contrast, UniDecouple combines task-specialized and shared FFN experts with learnable token-level routing while retaining shared attention, mitigating cross-task interference and preserving existing capabilities during continual MMU--T2I adaptation.}

\section{The Proposed UniDecouple}

\begin{figure}[H]
    \centering
    \includegraphics[width=1\linewidth,trim={1cm 2cm 2.5cm 3cm},clip]{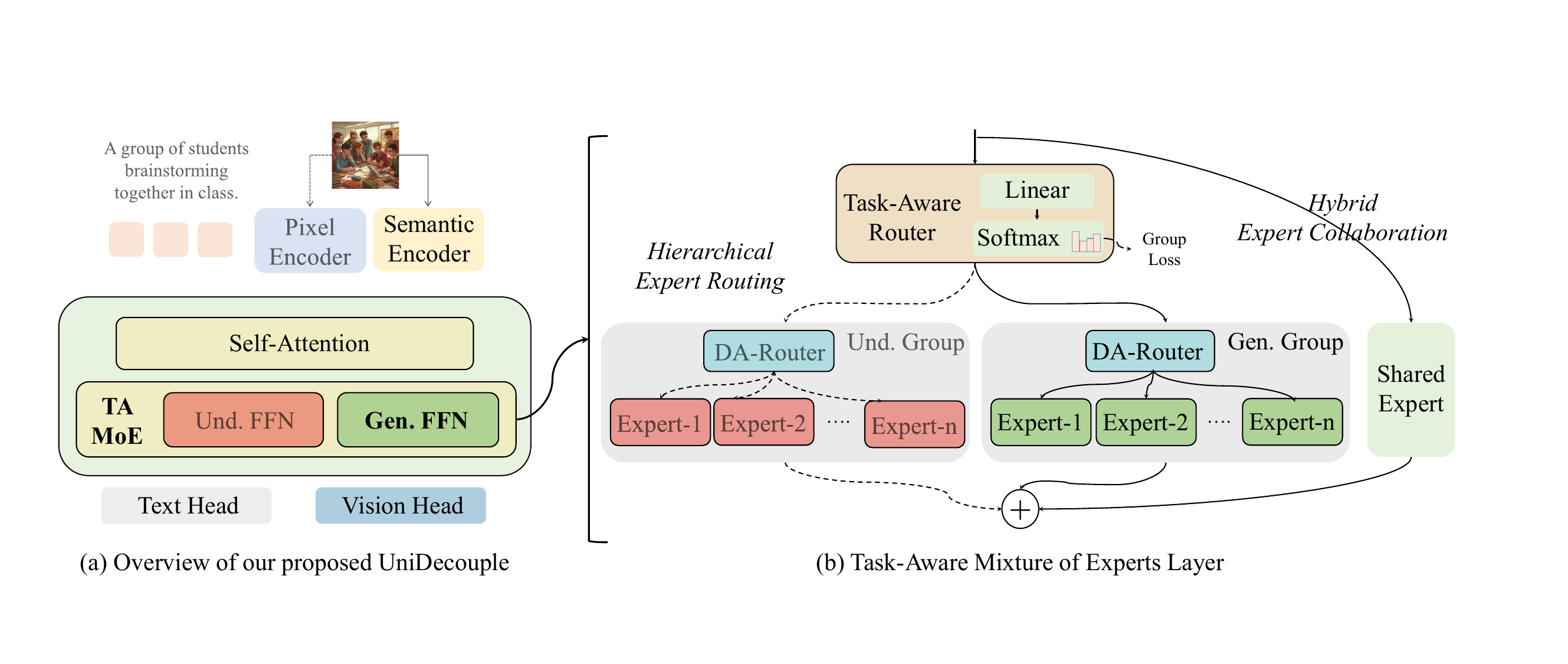}
        \caption{a) \textbf{Overview of UniDecouple}. ``Und." and ``Gen." denote MMU and T2I. b) \textbf{Task-Aware Mixture-of-Experts Layer}. Hierarchical Expert Routing consists of a task-aware router and a dynamic-assignment (DA) router, while Hybrid expert collaboration refers to integration of task-specific experts and shared experts.}
    \label{fig:Architecture}
\end{figure}

In this section, we first \revisiontext{theoretically characterize TOC} in Sec.~\ref{subsec:conflict}, \revisiontext{revealing task-level optimization interference and token-level complementarity between MMU and T2I}. Building on this foundation, we introduce \revisiontext{the core component of UniDecouple, the TA-MoE layer, which enables selective FFN decoupling and adaptive expert routing while preserving cross-task knowledge sharing} (Sec.~\ref{tamoe}). Furthermore, \revisiontext{we present a two-stage training strategy to balance task specialization and knowledge preservation during continual adaptation} in Sec.~\ref{train}.

\subsection{Task Objective Conflict}
\label{subsec:conflict}

\begin{revision}
First, we define the MMU and T2I objectives as autoregressive token-level cross-entropy losses:
\begin{equation}
\mathcal L_k(\theta)
=
\mathbb E_{x\sim\mathcal D_k}\!\left[
\sum_{t\in\mathcal T_k}\ell_t^{(k)}
\right],
\qquad
\ell_t^{(k)}
=
-\sum_{v\in\mathcal V_k}
y_{t,v}\log p_\theta(v\mid x_{<t}),
\quad k\in\{U,G\},
\end{equation}
where $\mathcal T_k$ denotes supervised positions, $\mathcal V_k$ is the output vocabulary, and $y_{t,v}=\mathbf 1[v=x_t]$ is the one-hot target. MMU predicts text tokens, while T2I predicts discrete image tokens. Below, we analyze empirical losses on fixed batches. For notational simplicity, $t\in\mathcal T_k$ indexes all supervised token occurrences across the fixed batch for task $k$, with batch and token normalization weights absorbed into $\ell_t^{(k)}$. Thus, $\mathcal L_k(\theta)=\sum_{t\in\mathcal T_k}\ell_t^{(k)}(\theta)$ in the following fixed-batch analysis. Sequential MMU and T2I updates give
\begin{equation}
\theta_1=\theta_0-\eta\nabla_\theta\mathcal L_U(\theta_0),
\qquad
\theta_2=\theta_1-\eta g_G,
\qquad
g_k=\nabla_\theta\mathcal L_k(\theta_1).
\end{equation}
Under local smoothness, a first-order Taylor expansion yields
\begin{equation}
\Delta\mathcal L_U
=
\mathcal L_U(\theta_2)-\mathcal L_U(\theta_1)
=
-\eta\langle g_U,g_G\rangle+\mathcal O(\eta^2).
\label{eq:generic-interference}
\end{equation}
Thus, $\langle g_U,g_G\rangle<0$ predicts \textbf{task-level} negative transfer for sufficiently small $\eta$. This condition is generic to sequential multi-task optimization and is not specific to AR.
To localize this interference, let
$g_t^{(k)}=\nabla_\theta\ell_t^{(k)}|_{\theta=\theta_1}$.
Then
\begin{equation}
g_U=\sum_{t\in\mathcal T_U}g_t^{(U)},
\qquad
g_G=\sum_{t'\in\mathcal T_G}g_{t'}^{(G)},
\qquad
A_{t,t'}^{UG}
=
\left\langle g_t^{(U)},g_{t'}^{(G)}\right\rangle,
\end{equation}
and
\begin{equation}
\langle g_U,g_G\rangle
=
\sum_{t\in\mathcal T_U}\sum_{t'\in\mathcal T_G}
A_{t,t'}^{UG}
=
\mathbf 1_U^\top\mathbf A^{UG}\mathbf 1_G.
\label{eq:token-pair}
\end{equation}
Define the negative and positive interaction masses as
\begin{equation}
C_-=\sum_{A_{t,t'}^{UG}<0}|A_{t,t'}^{UG}|,
\qquad
C_+=\sum_{A_{t,t'}^{UG}>0}A_{t,t'}^{UG}.
\end{equation}
Consequently,
\begin{equation}
\Delta\mathcal L_U
=
\eta(C_--C_+)+\mathcal O(\eta^2).
\label{eq:conflict-mass}
\end{equation}
For sufficiently small $\eta$, local negative transfer therefore occurs when conflicting \textbf{token-pair interactions} outweigh complementary ones.
Next, consider the causal mask under shifted next-token inputs, where the input at position $t$ is the previous token $x_{t-1}$ (or a start token at the first position):
\begin{equation}
M_{t,s}
=
\begin{cases}
0, & s\le t,\\
-\infty, & s>t.
\end{cases}
\end{equation}
Let $h_t^{(k)}$ be the final hidden representation for predicting token $t$, and $\theta_A$ the shared attention parameters. At $\theta_1$, define
\begin{equation}
z_t^{(k)}(M_k)
=
\frac{\partial\ell_t^{(k)}}{\partial h_t^{(k)}},
\qquad
J_t^{(k)}(M_k)
=
\frac{\partial h_t^{(k)}(M_k)}{\partial\theta_A}.
\end{equation}
The gradient with respect to shared attention parameters is
\begin{equation}
g_t^{(k),\mathrm{Attn}}
=
{J_t^{(k)}(M_k)}^\top z_t^{(k)}(M_k).
\end{equation}
Hence,
\begin{equation}
A_{t,t'}^{UG,\mathrm{Attn}}(M_U,M_G)
=
{z_t^{(U)}(M_U)}^\top
K_{\mathrm{Attn}}(t,t';M_U,M_G)
z_{t'}^{(G)}(M_G),
\label{eq:masked-interaction}
\end{equation}
where
\begin{equation}
K_{\mathrm{Attn}}(t,t';M_U,M_G)
=
J_t^{(U)}(M_U)
{J_{t'}^{(G)}(M_G)}^\top.
\label{eq:masked-kernel}
\end{equation}
Masking changes the accessible computational paths and thus both this cross-task Jacobian kernel and the loss gradients $z_t^{(k)}$.

Holding parameters, batches, targets, and loss normalization fixed, we compare the causal mask $M^{\mathrm C}$ defined above with a full-attention mask $M^{\mathrm F}$, whose entries are zero over valid token positions:
\begin{equation}
\Delta A_{t,t'}^{\mathrm{mask}}
=
A_{t,t'}^{UG,\mathrm{Attn}}(M^{\mathrm C},M^{\mathrm C})
-
A_{t,t'}^{UG,\mathrm{Attn}}(M^{\mathrm F},M^{\mathrm F}).
\label{eq:mask-effect}
\end{equation}
Negative values indicate reduced gradient alignment under causal masking; positive values indicate increased alignment. Full attention is only a diagnostic intervention, as it may expose future tokens. Sec.~\ref{confirm} will provide relevant empirical evidence.
\end{revision}

\subsection{Task-Aware Mixture of Experts Layer}
\label{tamoe}

\revisiontext{Motivated by our TOC analysis}, to better accommodate task objective conflict of MMU and T2I \revisiontext{during continual adaptation}, we introduce a modular design based on a MoE framework.
\revisiontext{Unlike MoT-style architectures that further separate attention modules, our MoE design retains shared attention to preserve pretrained contextual knowledge while specializing FFNs to mitigate task interference. This selective decoupling is better suited to continual learning, as it enables task-specific adaptation with limited architectural modifications and training overhead.}
Specifically, the experts are divided into two groups, each tailored to understanding and generation tasks, respectively.
To enhance task-specific specialization, we propose a Hierarchical Expert Routing mechanism, which comprises a task-aware router and a conventional dynamic assignment router. This mechanism enables more accurate routing of task-relevant tokens to the appropriate expert. Furthermore, to ensure coherence and synergy across task objectives, we incorporate a shared expert independent of the task-specific groups.
As a core part of the Hybrid Expert Collaboration mechanism, the shared expert facilitates cross-task information exchange and helps maintain a balanced representation between MMU and T2I tasks.

\subsubsection{Hierarchical Expert Routing}

To achieve fine-grained expert selection tailored to different task objectives, we introduce Hierarchical Expert Routing, consisting of a Task-Aware Router and a Dynamic-Assignment Router.

\noindent\textbf{Task-Aware Router.}
\begin{revision}
Given a token representation $x\in\mathbb R^d$, we employ a learnable, task-supervised hard router to select between the understanding and generation expert groups. The router first predicts token-dependent group probabilities:
\begin{equation}
\boldsymbol{\pi}_{\phi}(x)
=
\operatorname{Softmax}(W_r x+b_r),
\qquad
\hat g(x)
=
\arg\max_{g\in\{U,G\}}\pi_{\phi,g}(x),
\end{equation}
where $\phi=\{W_r,b_r\}$ denotes the trainable router parameters. The routing decision is predicted from each token representation rather than directly assigned using the sample-level task identity.
\end{revision}

\noindent\textbf{Dynamic-Assignment Router.}
\begin{revision}
Once the expert group $\hat g(x)$ is selected, we compute group-wise routing probabilities and select the top-$k_{\mathrm{exp}}$ experts within that group:
\begin{equation}
q_{g,i}(x)
=
\frac{\exp\!\left((\mathcal W_e^{(g)}x)_i\right)}
{\sum_{j=1}^{e}\exp\!\left((\mathcal W_e^{(g)}x)_j\right)},
\qquad i=1,\ldots,e.
\end{equation}
\begin{equation}
\mathcal S_{\hat g(x)}(x)
=
\operatorname{TopKIndices}\!\left(
\{q_{\hat g(x),i}(x)\}_{i=1}^{e},k_{\mathrm{exp}}
\right),
\qquad k_{\mathrm{exp}}=1.
\end{equation}
Here, $\mathcal W_e^{(g)}\in\mathbb R^{e\times d}$ is the group-specific expert scoring matrix, and $e$ is the number of experts in each group. $\operatorname{TopKIndices}$ returns the indices of the $k_{\mathrm{exp}}$ largest routing probabilities. The selected probabilities are retained without renormalization over the selected set.
\end{revision}

\subsubsection{Hybrid Expert Collaboration}

\begin{revision}
The selected expert outputs are weighted by their group-wise routing probabilities:
\begin{equation}
w_{g,i}(x)=q_{g,i}(x),
\end{equation}
\begin{equation}
h_{\mathrm{expert}}(x)
=
\sum_{i\in\mathcal S_{\hat g(x)}(x)}
w_{\hat g(x),i}(x)\,
\operatorname{Expert}_{\hat g(x),i}(x).
\end{equation}
Here, $\operatorname{Expert}_{g,i}$ denotes the $i$-th expert in group $g$.
To leverage both task-specific and global knowledge, we integrate the output of a shared expert:
\begin{equation}
h_{\mathrm{TA}}(x)
=
h_{\mathrm{expert}}(x)
+\alpha\,\operatorname{SharedExpert}(x),
\end{equation}
where $h_{\mathrm{TA}}(x)$ denotes the output of the TA-MoE layer and $\alpha$ is a fixed hyperparameter set to $0.2$ based on repeated empirical testing to balance the contributions.
\end{revision}

\noindent\textbf{Loss Function.}
\begin{revision}
The overall training objective includes the task-specific losses and a group-supervision loss. We average the group-supervision loss over the token positions and TA-MoE layers included in routing supervision:
\begin{equation}
\mathcal L_{\mathrm{group}}
=
-\frac{1}{|\mathcal R|}
\sum_{(b,t,l)\in\mathcal R}
\log\pi_{\phi_l,g_b^*}\!\left(x_{b,t}^{(l)}\right).
\end{equation}
Here, $\mathcal R$ is the set of sample-token-layer tuples included in routing supervision, $x_{b,t}^{(l)}$ is the input representation of token $t$ from sample $b$ at TA-MoE layer $l$, $\phi_l$ denotes the task-aware router parameters at layer $l$, and $g_b^*\in\{U,G\}$ is the ground-truth task label of sample $b$. Layer indices are omitted in the preceding routing equations for clarity.
The overall objective is
\begin{equation}
\mathcal L_{\mathrm{total}}(\Theta)
=
\lambda_U\mathcal L_U(\Theta)
+\lambda_G\mathcal L_G(\Theta)
+\gamma\mathcal L_{\mathrm{group}}(\Theta).
\end{equation}
Here, $\Theta$ denotes the parameters of the complete UniDecouple model, including the routers and experts. $\mathcal L_U$ and $\mathcal L_G$ are the autoregressive token-level cross-entropy objectives defined in Sec.~\ref{subsec:conflict}, evaluated using the complete model. The coefficients $\lambda_U$, $\lambda_G$, and $\gamma$ control the contributions of the understanding, generation, and routing-supervision losses, respectively.
\end{revision}

\subsection{Training Strategy}
\label{train}

To fully leverage the unique capabilities of different experts and enhance the generalization ability of the overall framework, we propose a Two-stage Training strategy, as illustrated in Fig.~\ref{fig:training}. In the following paragraphs, we detail objectives and provide a comprehensive explanation of each training stage.

\noindent\textbf{Stage 1: Training Task-Specific Experts.}
\begin{revision}
This stage derives task-specific experts through separate training
for each task. For each task $k\in\{U,G\}$, we optimize the
task-specific model using the autoregressive token-level
cross-entropy objective:
\begin{equation}
\mathcal L_{\mathrm{stage1}}^{(k)}(\Theta_k)
=
-\mathbb E_{x\sim\mathcal D_k}\!\left[
\sum_{t\in\mathcal T_k}
\sum_{v\in\mathcal V_k}
y_{t,v}\log p_{\Theta_k}(v\mid x_{<t})
\right],
\qquad k\in\{U,G\}.
\end{equation}
Here, $\Theta_k$ denotes the parameters of the task-specific
model, $\mathcal T_k$ denotes the supervised positions in each
sequence, and $\mathcal V_k$ is the output vocabulary for task $k$.
The target distribution is one-hot, with
$y_{t,v}=\mathbf 1[v=x_t]$, and
$p_{\Theta_k}(v\mid x_{<t})$ is the model-predicted probability
of token $v$ given the preceding context. Only the FFN parameters are optimized in this stage.

\begin{figure}[H]
    \centering
    \includegraphics[width=0.8\linewidth,trim={14cm 3cm 13cm 3.5cm},clip]{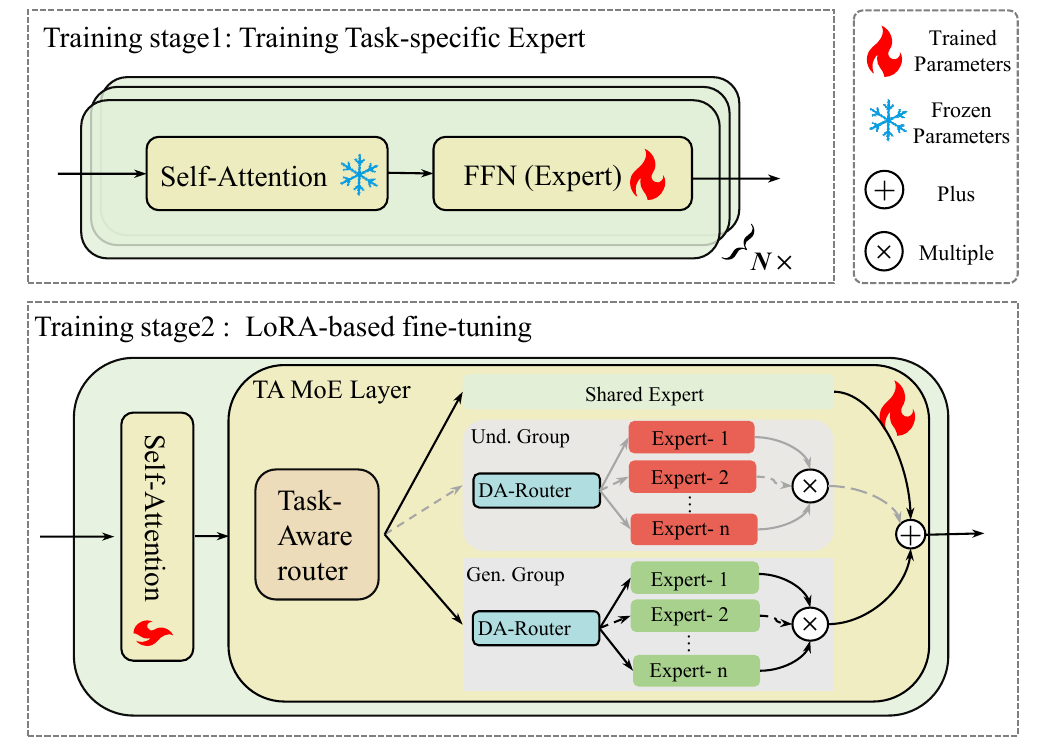}
    \caption{Illustration of the proposed two-stage training framework. Stage 1 derives highly specialized task-specific experts by freezing the shared self-attention layers and optimizing only FFN parameters under the \revisiontext{autoregressive cross-entropy objective}. Stage 2 replaces conventional FFNs with the TA-MoE layer, incorporating the pre-trained experts and performing end-to-end LoRA-based fine-tuning. This design enables effective task decoupling while maintaining cross-task knowledge sharing and parameter-efficient adaptation.}
    \label{fig:training}
\end{figure}

Let $\Theta_k=(\mathcal W_{\mathrm{expert},k},\Psi_k)$, where $\mathcal W_{\mathrm{expert},k}$ contains the FFN parameters and $\Psi_k$ contains all remaining parameters, including the self-attention parameters. The parameter updates are
\begin{equation}
\begin{aligned}
\mathcal W_{\mathrm{expert},k}^{(n+1)}
&=
\mathcal W_{\mathrm{expert},k}^{(n)}
-\eta
\left.
\nabla_{\mathcal W_{\mathrm{expert},k}}
\mathcal L_{\mathrm{stage1}}^{(k)}(\Theta_k)
\right|_{\Theta_k=\Theta_k^{(n)}},
\\
\Psi_k^{(n+1)}&=\Psi_k^{(n)}.
\end{aligned}
\end{equation}
Here, $n$ denotes the optimization step and $\eta$ is the learning rate. All non-FFN parameters remain fixed while the FFN parameters are adapted to the corresponding task.
\end{revision}

\noindent\textbf{Stage 2: LoRA-Based Fine-Tuning of UniDecouple.}
\begin{revision}
We replace the conventional FFNs with TA-MoE layers and initialize the task-specific experts using the weights obtained in Stage 1. The output of each TA-MoE layer is
\begin{equation}
h_{\mathrm{TA}}(x)
=
\sum_{i\in\mathcal S_{\hat g(x)}(x)}
w_{\hat g(x),i}(x)\,
\operatorname{Expert}_{\hat g(x),i}(x)
+
\alpha\,\operatorname{SharedExpert}(x),
\end{equation}
where $\hat g(x)$ is the group selected by the learnable task-aware router, $\mathcal S_{\hat g(x)}(x)$ is the set of experts selected by the dynamic-assignment router within that group, and $w_{\hat g(x),i}(x)=q_{\hat g(x),i}(x)$ is the group-wise routing probability defined above, retained without renormalization over the selected set. Experts in the unselected group do not contribute to this token's output.
\end{revision}

\begin{revision}
We perform end-to-end LoRA-based fine-tuning with rank $r=16$. For each weight matrix adapted with LoRA, we parameterize its update as
\begin{equation}
\mathcal W'
=\mathcal W+\Delta\mathcal W
=\mathcal W+s_{\mathrm{LoRA}}\mathcal B\mathcal A,
\qquad
\mathcal W\in\mathbb R^{d_{\mathrm{out}}\times d_{\mathrm{in}}},
\end{equation}
where
\begin{equation}
\mathcal B\in\mathbb R^{d_{\mathrm{out}}\times r},
\qquad
\mathcal A\in\mathbb R^{r\times d_{\mathrm{in}}},
\qquad r=16.
\end{equation}
The coefficient $s_{\mathrm{LoRA}}$ denotes the scaling factor used in the implementation; setting $s_{\mathrm{LoRA}}=1$ recovers the unscaled update. For each adapted matrix, the pretrained weight $\mathcal W$ remains fixed, while $\mathcal A$ and $\mathcal B$ are trainable. We optimize the inserted LoRA adapters and the trainable routing parameters using $\mathcal L_{\mathrm{total}}$ through end-to-end backpropagation. The task-aware routers receive supervision from $\mathcal L_{\mathrm{group}}$, while the dynamic-assignment routers receive task-loss gradients through the retained group-wise routing probabilities. This enables task-specific adaptation and cross-task knowledge sharing with parameter-efficient updates.
\end{revision}

\section{Experiments}

In this section, we present comprehensive experiments for our proposed method. First, we describe the experimental setup, including the model architecture and training configurations. Next, we verify the existence of task objective conflicts (TOC) through metrics such as loss and evaluate the model on various understanding and generation benchmarks. We then provide qualitative results to illustrate the strengths of our approach. Finally, we analyze the utilization of both task-specific and shared experts.

\subsection{Experimental Setups}

\noindent\textbf{Model Architecture.} We adopt JanusPro as the base model, whose language component supports a maximum sequence length of 4096. For multimodal understanding, we use SigLIP-Large-Patch16-384~\cite{zhai2023sigmoid} as the image encoder. For image generation, the encoder utilizes a codebook of size 16,384 and downsamples inputs by a factor of 16. Both understanding and generation adapters are implemented as two-layer MLPs. 

\noindent\textbf{Training Configurations.} For the training setup, we keep the vision tower frozen in all stages since it is sufficiently pretrained by~\cite{chen2025janus}. Task-specific experts are trained on ShareGPT4V~\cite{chen2024sharegpt4v} for understanding and MidJourney~\cite{vivym2023midjourney} for generation, while the LoRA fine-tuning stage uses a mixed dataset of both. In the MoE design, each task group is assigned two experts with Top-$k$ routing, and one shared expert is added to enhance generalization. The hyperparameters are as follows: in Stage 1, batch size of 32, learning rate $1.0 \times 10^{-4}$, and 200 training steps; in Stage 2, batch size of 32, learning rate $2.0 \times 10^{-5}$, and 400 training steps. Both stages use the AdamW optimizer ($\beta_1=0.9,\ \beta_2=0.95$), cosine learning rate scheduler, and weight decay of 0.0. The main loss coefficients are set to $\lambda_1=0.3$, $\lambda_2=0.3$, and $\alpha=0.2$.

\subsection{Existence of Conflict}
\label{confirm}

\begin{figure*}[t]
    \begin{revision}
    \centering
    \begin{minipage}[t]{0.48\textwidth}
        \centering
        \includegraphics[width=\linewidth]{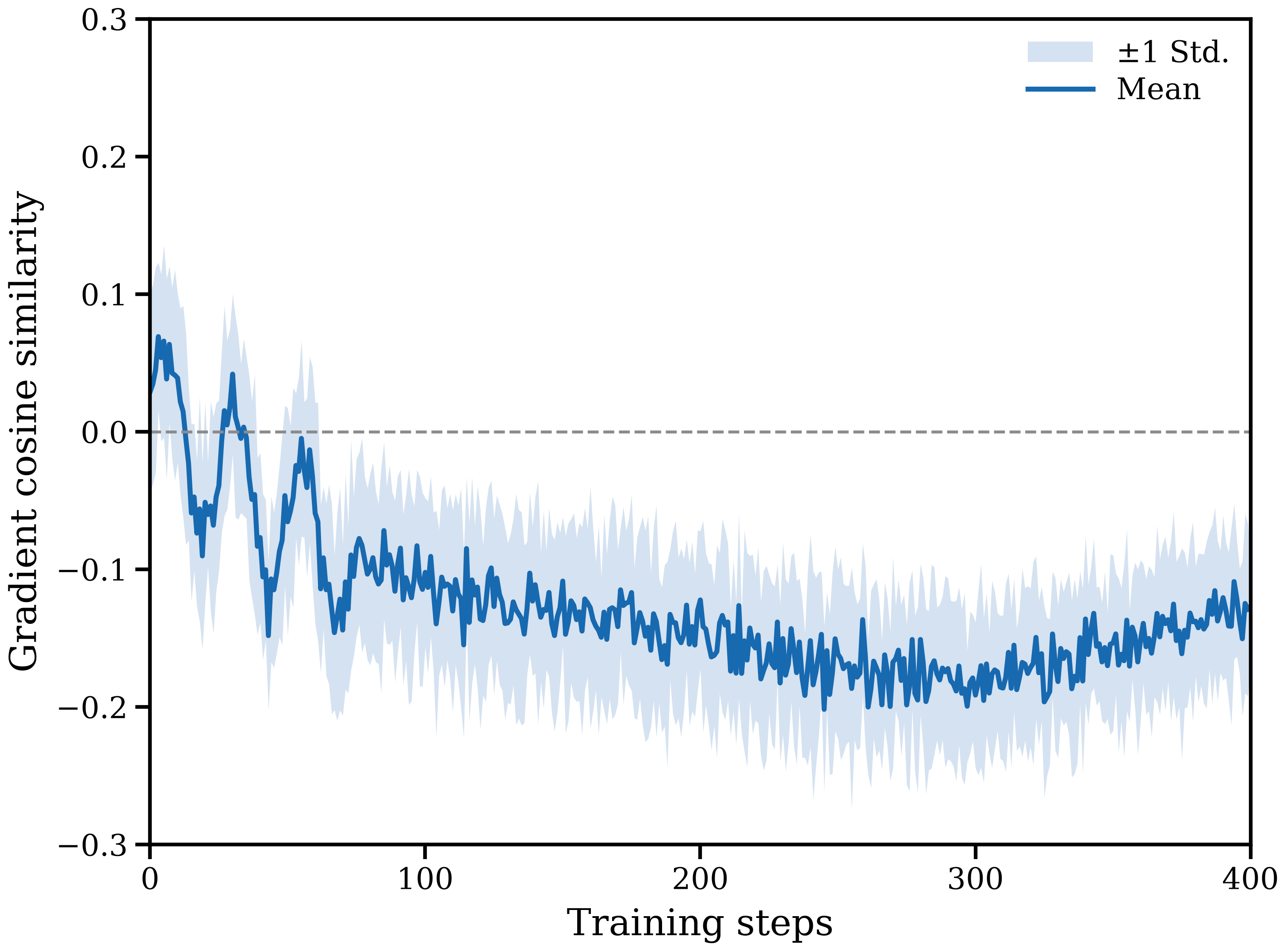}
        \par\small (a) Task-level gradient cosine similarity.
    \end{minipage}
    \hfill
    \begin{minipage}[t]{0.48\textwidth}
        \centering
        \includegraphics[width=\linewidth]{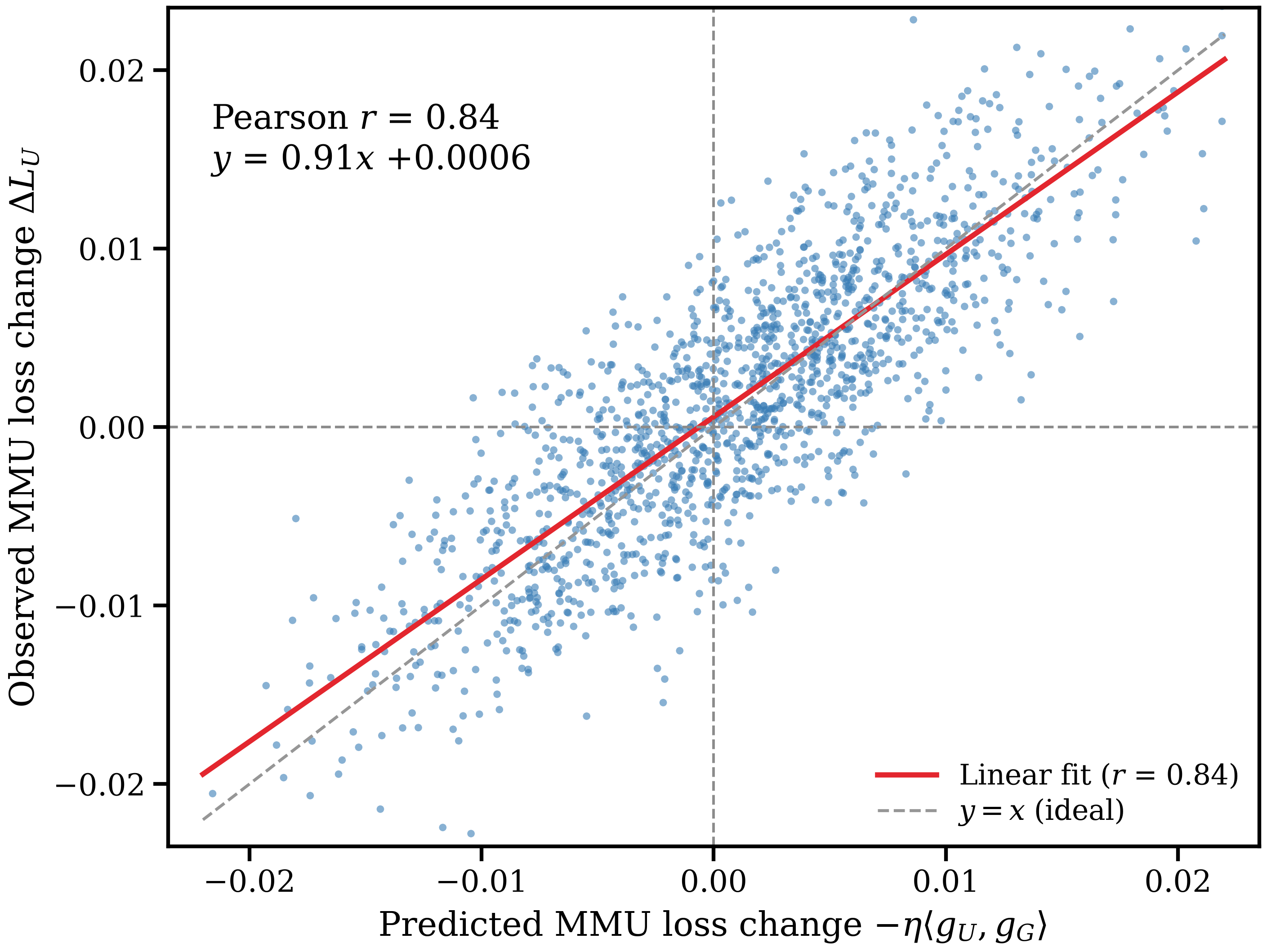}
        \par\small (b) Predicted and observed MMU loss changes.
    \end{minipage}
    \captionsetup{labelfont={color=black},textfont={color=black}}
    \caption{Direct validation of task-level gradient conflict between MMU and T2I. (a) Gradient cosine similarity during joint training; the solid curve and shaded region indicate the mean and $\pm1$ standard deviation over independent runs. Negative values indicate gradient conflict. (b) First-order prediction $-\eta\langle g_U,g_G\rangle$ versus the observed MMU loss change after a T2I gradient step.}
    \label{fig:toc_grad_validation}
    \end{revision}
\end{figure*}

\begin{revision}
We investigate TOC during joint MMU--T2I training using the baseline based on~\cite{chen2025janus}. Following the analysis in Sec.~\ref{subsec:conflict}, we examine four complementary aspects: task-level gradient interference, token-level interaction localization, the effect of causal masking and downstream benchmark evaluation. We additionally reserve training loss dynamics as supporting observations.

\noindent\textbf{Task-level Gradient Interference.}
According to Eq.~\eqref{eq:generic-interference}, the first-order change in the MMU loss under a T2I gradient step is
\begin{equation}
\Delta\mathcal L_U
=\mathcal L_U(\theta-\eta g_G)-\mathcal L_U(\theta)
\approx-\eta\langle g_U,g_G\rangle.
\label{eq:toc-first-order}
\end{equation}
Thus, a negative gradient inner product predicts a local increase in the MMU loss for sufficiently small $\eta$. During joint training, we periodically sample one MMU mini-batch and one T2I mini-batch and measure both their raw gradient inner product and cosine similarity:
\begin{equation}
\cos(g_U,g_G)
=\frac{\langle g_U,g_G\rangle}{\|g_U\|_2\|g_G\|_2}.
\label{eq:toc-task-cosine}
\end{equation}
As shown in Fig.~\ref{fig:toc_grad_validation}(a), the gradients predominantly negative after the initial training phase, with an average cosine similarity of $-0.1327\pm0.0219$ and an average raw inner product of $-1.2664\pm0.2137$ across independent runs. Both measurements support the gradient-conflict condition in Sec.~\ref{subsec:conflict}.

We further compare the observed MMU loss change under the gradient step in Eq.~\eqref{eq:toc-first-order} with its first-order prediction $-\eta\langle g_U,g_G\rangle$. Fig.~\ref{fig:toc_grad_validation}(b) shows a Pearson correlation of $0.84$ between the predicted and observed changes, connecting negative gradient alignment to measurable local cross-task interference.
\end{revision}

\begin{figure*}[t]
    
    \centering
    \begin{subfigure}[t]{0.48\textwidth}
        \centering
        \vspace{0pt}
        \includegraphics[width=\linewidth]{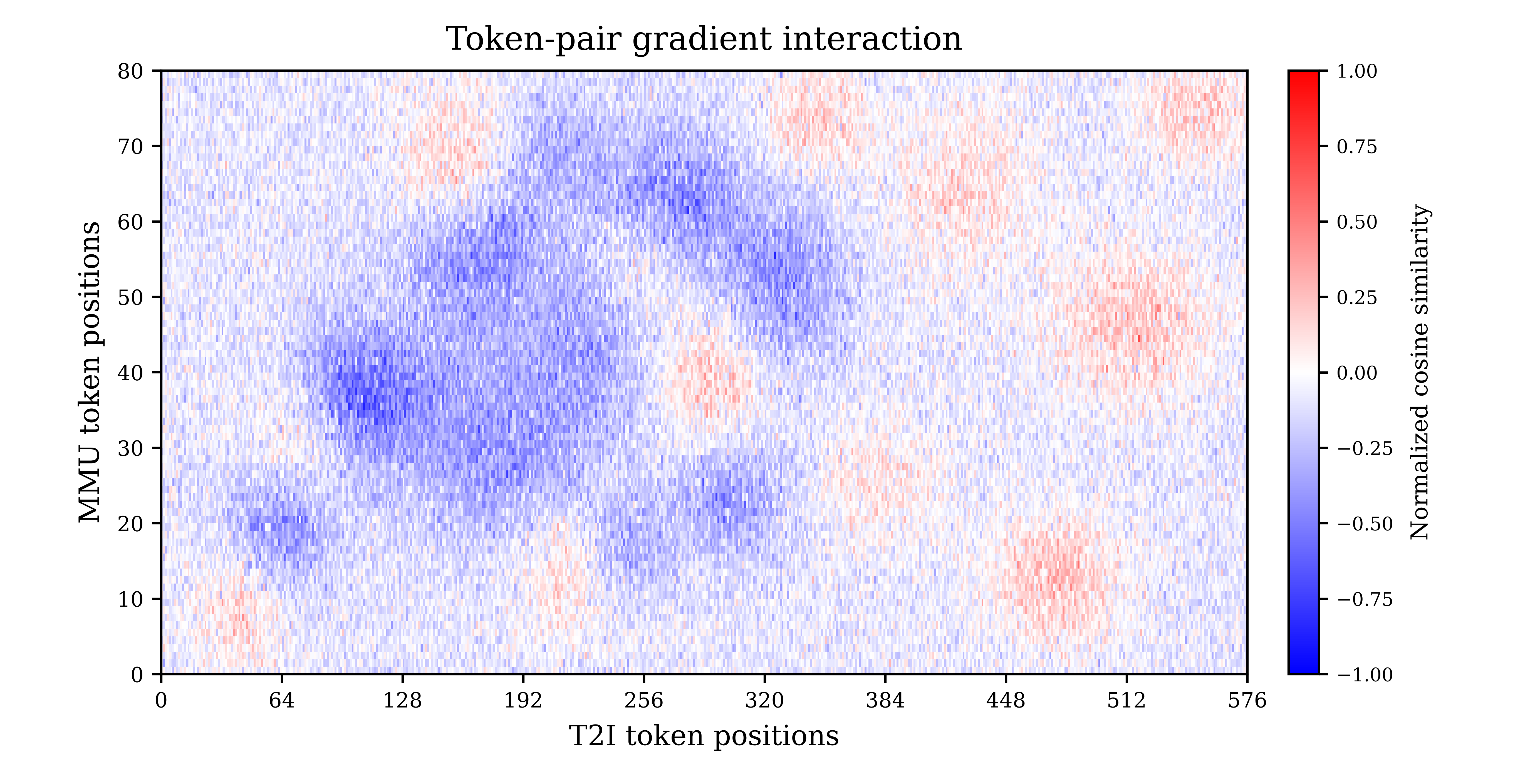}
        \captionsetup{labelfont={color=black},textfont={color=black}}
        \caption{Token-level gradient interactions.}
        \label{fig:toc_token_heatmap}
    \end{subfigure}
    \hfill
    \begin{subfigure}[t]{0.48\textwidth}
        \centering
        \vspace{0pt}
        \includegraphics[width=\linewidth,trim={3.5cm 1cm 3.5cm 1cm},clip]{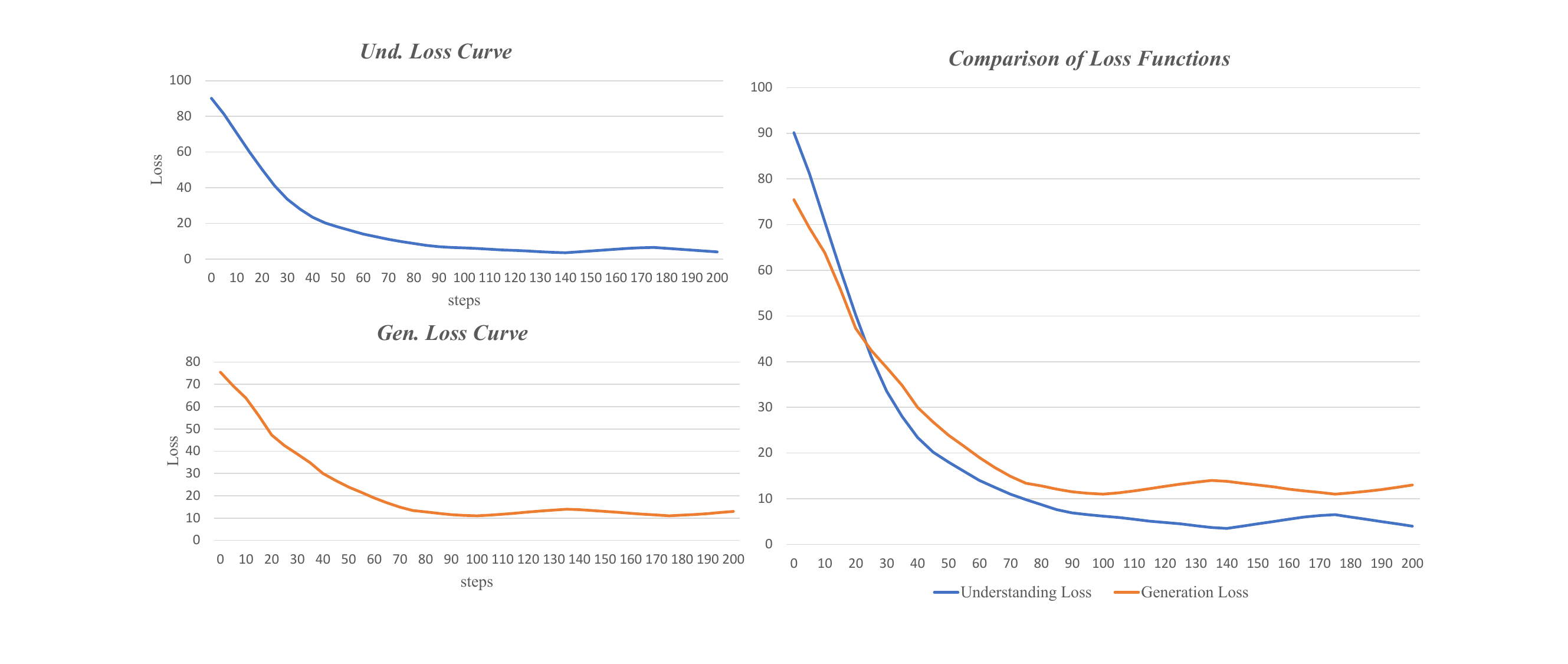}
        \caption{MMU and T2I training loss dynamics.}
        \label{fig:lossfunc}
    \end{subfigure}

    \par\medskip
    \begin{subfigure}[t]{\textwidth}
        \centering
        \includegraphics[width=\linewidth]{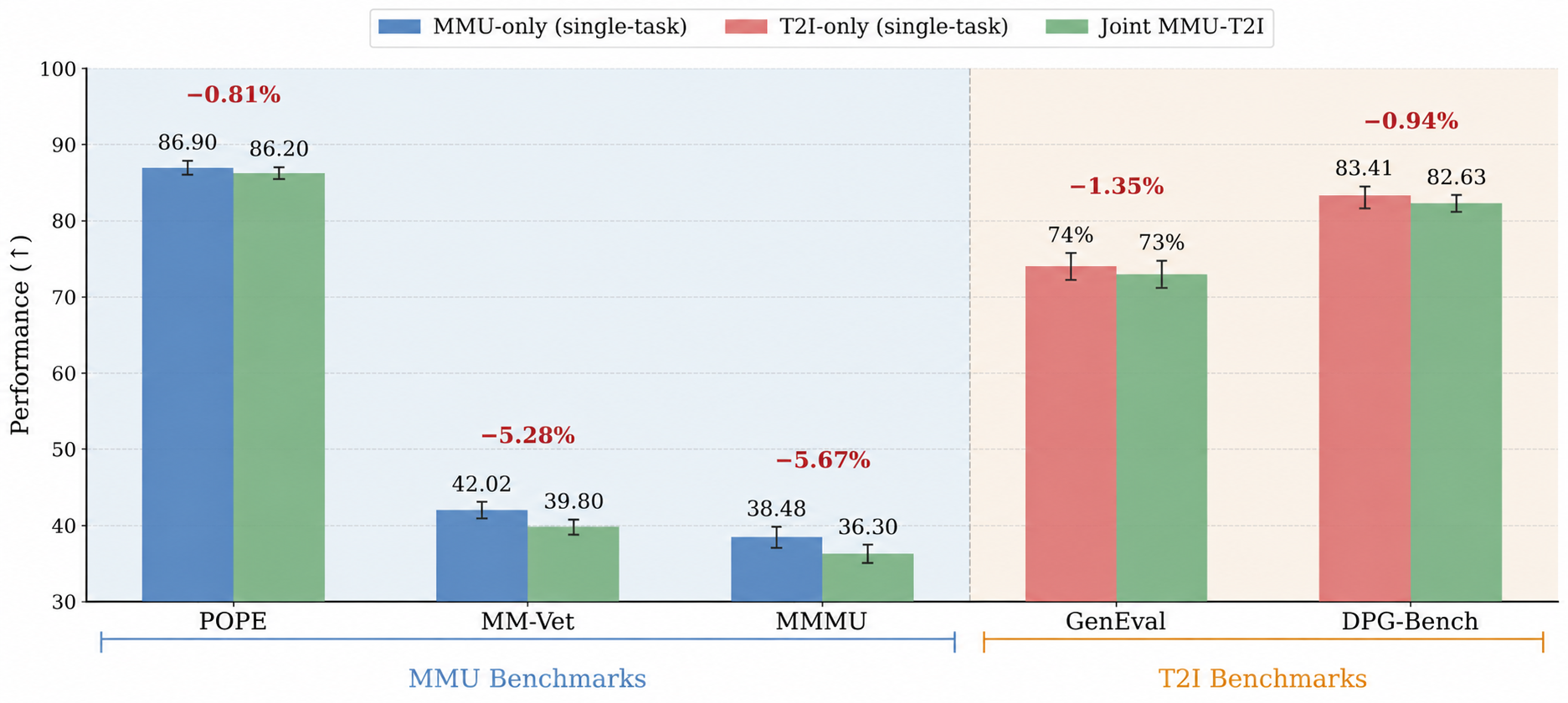}
        \captionsetup{labelfont={color=black},textfont={color=black}}
        \caption{Downstream benchmark comparison.}
        \label{fig:toc_benchmark_comparison}
    \end{subfigure}
    \captionsetup{labelfont={color=black},textfont={color=black}}
    \caption{Cross-task interference between MMU and T2I. (a) Mean token-pair gradient cosine similarity across normalized prediction-position bins, using the same samples as the task-level analysis. Values range from $-1$ to $1$; negative and positive values indicate conflict and complementarity, respectively. (b) Autoregressive cross-entropy loss curves showing cross-task trade-offs. (c) Single-task versus joint-training performance: blue, red, and green denote MMU-only, T2I-only, and joint training, respectively. Red annotations indicate relative changes defined in Eq.~\eqref{eq:toc-benchmark-change}. GenEval scores are scaled by $100$; higher is better on all benchmarks.}
    \label{fig:toc_interference_analysis}
\end{figure*}

\begin{revision}

\noindent\textbf{Token-level Interaction Localization.}
To identify where task-level interference arises, we use the same MMU--T2I samples as in the task-level measurements above. As defined in Sec.~\ref{subsec:conflict}, the task gradients decompose as $g_U=\sum_{t\in\mathcal T_U}g_t^{(U)}$ and $g_G=\sum_{t'\in\mathcal T_G}g_{t'}^{(G)}$, with loss-averaging weights absorbed into the token contributions. The raw interaction is $A_{t,t'}^{UG}=\langle g_t^{(U)},g_{t'}^{(G)}\rangle$. For visualization, we compute the cosine similarity for every MMU--T2I token pair with nonzero gradients:
\begin{equation}
\widetilde A_{t,t'}^{UG}
=\frac{\langle g_t^{(U)},g_{t'}^{(G)}\rangle}
{\|g_t^{(U)}\|_2\|g_{t'}^{(G)}\|_2}
\in[-1,1].
\label{eq:toc-token-cosine}
\end{equation}
The raw inner product satisfies $A_{t,t'}^{UG}=\|g_t^{(U)}\|_2\|g_{t'}^{(G)}\|_2\widetilde A_{t,t'}^{UG}$; it therefore has the same sign as the cosine similarity but additionally accounts for both gradient magnitudes.

Since sequence lengths vary across tasks and samples, we map token positions to normalized relative locations,
\begin{equation}
\bar t=\frac{t}{|\mathcal T_U|},
\qquad
\bar t'=\frac{t'}{|\mathcal T_G|},
\end{equation}
and average token-pair cosine similarities within a fixed $B_U\times B_G$ grid over normalized position space. The resulting heatmap remains within $[-1,1]$ and is comparable across samples of different sequence lengths. Fig.~\ref{fig:toc_token_heatmap} shows that conflicting (negative) and complementary (positive) interactions are non-uniformly distributed across prediction positions.

Before spatial aggregation, we summarize the raw interactions by their negative and positive fractions:
\begin{equation}
p_-=
\frac{\sum_{t,t'}\mathbf 1[A_{t,t'}^{UG}<0]}{|\mathcal T_U||\mathcal T_G|},
\qquad
p_+=
\frac{\sum_{t,t'}\mathbf 1[A_{t,t'}^{UG}>0]}{|\mathcal T_U||\mathcal T_G|}.
\end{equation}
Using the raw interaction masses $C_-$ and $C_+$ defined in Sec.~\ref{subsec:conflict}, Eq.~\eqref{eq:token-pair} gives
\begin{equation}
\langle g_U,g_G\rangle=\sum_{t,t'}A_{t,t'}^{UG}=C_+-C_-.
\end{equation}
We obtain $p_-=0.6817$ and $p_+=0.3183$, with $C_-=2.6843$ and $C_+=1.4179$. Thus, $C_--C_+=1.2664$, consistent with the task-level inner product of $-1.2664$. These measurements localize the aggregate gradient conflict to token-pair interactions and show that negative interaction mass outweighs positive interaction mass.
\end{revision}

\begin{revision}

\noindent\textbf{Effect of Causal Masking.}
We test whether causal masking amplifies gradient conflict in the examined MMU--T2I setting. Holding model parameters, mini-batches, targets, and loss normalization fixed, we compare gradient interactions under causal attention $M^{\mathrm C}$ and full attention $M^{\mathrm F}$. The attention-specific token-level effect is characterized by $\Delta A_{t,t'}^{\mathrm{mask}}$ in Eq.~\eqref{eq:mask-effect}.

As summarized in Table~\ref{tab:toc_direct_validation}, the average task-level cosine similarity changes from $-0.0719\pm0.0187$ under full attention to $-0.1327\pm0.0219$ under causal attention, while $C_--C_+$ increases from $0.7124$ to $1.2664$. These results indicate stronger gradient interference under causal masking in this setting. Full attention serves only as a diagnostic intervention because it may expose future tokens; these observations do not establish conflict amplification as a universal property of causal masking.
\end{revision}

\begin{table*}[t]
    \centering
    \begin{revision}
    \captionsetup{labelfont={color=black},textfont={color=black}}
    \caption{Gradient conflict under full and causal attention. Statistics are averaged over MMU--T2I mini-batch pairs and independent runs. The task-level inner product and cosine similarity are reported as mean $\pm$ standard deviation across runs. $C_-$ and $C_+$ are the negative and positive raw token-pair interaction masses.}
    \label{tab:toc_direct_validation}
    \setlength{\tabcolsep}{4pt}
    \renewcommand{\arraystretch}{0.95}
    \small
    \begin{tabular}{lccccc}
    \toprule
    Attention & $\langle g_U,g_G\rangle$ & $\cos(g_U,g_G)$ & $C_-$ & $C_+$ & $C_--C_+$ \\
    \midrule
    Full & $-0.7124\pm0.1847$ & $-0.0719\pm0.0187$ & $2.1387$ & $1.4263$ & $0.7124$ \\
    Causal & $-1.2664\pm0.2137$ & $-0.1327\pm0.0219$ & $2.6843$ & $1.4179$ & $1.2664$ \\
    \bottomrule
    \end{tabular}
    \end{revision}
\end{table*}

\begin{revision}
\noindent\textbf{Downstream Benchmark Evaluation.}
To examine how cross-task interference manifests in downstream performance, we compare task-specific single-task training with joint MMU--T2I training over multiple random seeds. We report mean performance and standard deviation on three MMU benchmarks (POPE, MM-Vet, and MMMU) and two T2I benchmarks (GenEval and DPG-Bench). For each benchmark, we compute the absolute score change and the relative performance change as
\begin{equation}
\Delta S=S_{\mathrm{Joint}}-S_{\mathrm{Single}},
\qquad
\Delta S_{\mathrm{rel}}
=\frac{S_{\mathrm{Joint}}-S_{\mathrm{Single}}}{S_{\mathrm{Single}}}\times100\%.
\label{eq:toc-benchmark-change}
\end{equation}
Here, $S_{\mathrm{Single}}$ denotes the score of the corresponding MMU-only or T2I-only model, and $S_{\mathrm{Joint}}$ denotes that of the jointly trained model. All five benchmarks use higher-is-better metrics, so negative changes indicate degradation relative to single-task training.

Fig.~\ref{fig:toc_benchmark_comparison} shows lower mean performance under joint training across all five benchmarks, with different magnitudes of degradation. For MMU, POPE decreases from $86.90$ to $86.20$ ($-0.81\%$ relative), MM-Vet from $42.02$ to $39.80$ ($-5.28\%$), and MMMU from $38.48$ to $36.30$ ($-5.67\%$). For T2I, GenEval decreases from $0.74$ to $0.73$ ($-1.35\%$ relative), while DPG-Bench decreases from $83.41$ to $82.63$ ($-0.94\%$). GenEval scores are displayed as percentages in the figure; its absolute change is $-0.01$, equivalent to a decrease of one percentage point.

The larger relative drops on MM-Vet and MMMU indicate that the observed degradation varies across benchmarks. The error bars provide context for run-to-run variation, and the small mean differences on POPE and GenEval alone do not establish statistically significant degradation. Together with the direct gradient measurements, these results connect local optimization conflict to a broader pattern of lower downstream mean performance under joint training, without treating small benchmark changes as evidence of catastrophic forgetting.
\end{revision}

\begin{revision}
\noindent\textbf{Training Loss Dynamics.}
As a complementary observation, we track the MMU and T2I training losses, both of which use autoregressive token-level cross-entropy. Fig.~\ref{fig:lossfunc} shows intervals in which a decrease in one task's loss accompanies an increase in the other's. These patterns are consistent with cross-task interference, while the gradient measurements above provide direct evidence of local optimization conflict. Such local conflict and short-term loss changes alone do not establish catastrophic forgetting, which requires evidence of substantial and persistent degradation of previously acquired capabilities.
\end{revision}

\subsection{Quantitative Evaluation}
\label{compare}

\noindent\textbf{Multimodal Understanding (MMU).} To comprehensively assess multimodal understanding ability, we evaluate our model on several widely recognized benchmarks covering object perception, compositional reasoning, hallucination detection, and cross-domain generalization. 
Specifically, POPE~\cite{li2023evaluating} evaluates object hallucination in vision-language models; 
MME~\cite{mme} measures perception and cognition across multiple subtasks; 
MMBench (MMB)~\cite{liu2024mmbench} provides multi-level reasoning evaluation; 
SEED-Bench~\cite{li2024seed} tests multimodal knowledge and reasoning; 
GQA~\cite{hudson2019gqa} focuses on compositional visual question answering; 
MMMU~\cite{yue2024mmmu} examines expert-level multimodal reasoning across disciplines; 
and MM-Vet~\cite{yu2023mm} evaluates complex reasoning and instruction-following abilities.

\noindent\textbf{Text-to-Image Generation (T2I).} To evaluate text-to-image generation capability, we adopt challenging compositional and alignment-focused benchmarks. 
GenEval~\cite{ghosh2023geneval} measures compositional consistency and attribute binding between text prompts and generated images, while DPG-Bench~\cite{hu2024ella} evaluates fine-grained prompt following and dense semantic grounding performance. 
These benchmarks collectively assess image-text alignment, compositional reasoning, and controllable generation ability.

\noindent\textbf{Comparison Methods.} As shown in Table~\ref{compare-understanding} and Table~\ref{compare-generation1}, we categorize the models into three groups based on their capabilities. (i) \textit{Understanding Only}: This group includes models designed solely for MMU, such as the \textit{LLaVA series} (e.g., LLaVA~\cite{liu2023visual}, LLaVA-v1.5~\cite{liu2024improved}), the \textit{MobileVLM series} (e.g., MobileVLM~\cite{chu2023mobilevlm}, MobileVLM-V2~\cite{chu2024mobilevlm}), InstructBLIP~\cite{dai2023instructblipgeneralpurposevisionlanguagemodels}, Qwen-VL-Chat~\cite{bai2023qwen}, and Emu3-Chat~\cite{wang2024emu3}. (ii) \textit{Generation Only}: This group contains models that focus exclusively on image generation, including LlamaGen~\cite{sun2024autoregressive}, LDM~\cite{rombach2022high}, SDv1.5~\cite{rombach2022high}, PixArt-$\alpha$~\cite{chen2023pixart}, SDv2.1~\cite{rombach2022high}, DALL·E 2~\cite{ramesh2022hierarchical}, Emu3-Gen~\cite{wang2024emu3}, SDXL~\cite{podell2023sdxl}, DALL·E 3~\cite{betker2023improving}, and SD3-Medium~\cite{esser2024scaling}. (iii) \textit{Understanding and Generation}: This group comprises models capable of both MMU and T2I, such as DreamLLM~\cite{dong2023dreamllm}, LaViT~\cite{jin2023unified}, MetaMorph~\cite{tong2024metamorph}, NEXT-GPT~\cite{wu2024next}, Show-o-256~\cite{xie2024show}, SDHo-512~\cite{xie2024show}, D-DiT~\cite{li2024dual}, Gemini-Nano-1~\cite{team2023gemini}, ILLUME~\cite{wang2024illume}, TokenFlow-XL~\cite{qu2024tokenflow}, LWM~\cite{liu2024world}, VILA-U~\cite{wu2024vila}, Chameleon~\cite{team2024chameleon}, and the Janus-series models~\cite{wu2024janus,chen2025janus,ma2024janusflow}.

\begin{table*}[htbp]
  \centering
  \setlength{\tabcolsep}{4pt} 
  \small 
  \caption{Comparison with state-of-the-arts on MMU benchmarks. "Und." and "Gen." denote "understanding" and "generation", respectively. Model using external pretrained diffusion model are marked with †.}
    \resizebox{1\linewidth}{!}{%
  \begin{tabular}{llcccccccc}
    \hline
    \textbf{Type}           & \textbf{Model} & \textbf{Params}& \textbf{POPE↑} & \textbf{MME-P↑} & \textbf{MMB↑} & \textbf{SEED↑} & \textbf{GQA↑} & \textbf{MMMU↑} & \textbf{MM-Vet↑} \\
    \hline
    \multirow{12}{*}{Und. Only}& LlaVA-v1.5-Phi-1.5~\cite{xie2024show} & 1.3B & 84.1 & 1128.0 & - & - & 56.5 & \revisiontext{30.7} & - \\
    & MobileVLM~\cite{chu2023mobilevlm} & 1.4B & 84.5 & 1196.2 & 53.2 & - & 56.1 & - & - \\
    & MobileVLM-V2~\cite{chu2024mobilevlm} & 1.4B & 84.3 & 1302.8 & 57.7 & - & 59.3 & - & - \\
    & MobileVLM~\cite{chu2023mobilevlm} & 2.7B & 84.9 & 1288.9 & 59.6 & - & 59.0 & - & - \\
    & MobileVLM-V2~\cite{chu2024mobilevlm} & 2.7B & 84.7 & 1440.5 & 63.2 & - & 61.1 & - & - \\
    & LlaVA-Phi~\cite{zhu2024llavaphiefficientmultimodalassistant} & 2.7B & 85.0 & 1335.1 & 59.8 & - & - & - & 28.9 \\
    & LLaVA~\cite{liu2023visual} & 7B & 76.3 & 809.6 & 38.7 & 33.5 & - & - & 25.5 \\
    & LLaVA-v1.5~\cite{liu2024improved} & 7B & 85.9 & 1510.7 & 64.3 & 58.6 & 62.0 & 35.4 & 31.1 \\
    & InstructBLIP~\cite{dai2023instructblipgeneralpurposevisionlanguagemodels} & 7B & - & - & 36.0 & 53.4 & 49.2 & - & 26.2 \\
    & Qwen-VL-Chat~\cite{bai2023qwen} & 7B & - & 1487.5 & 60.6 & 58.2 & 57.5 & - & - \\
    & Emu3-Chat~\cite{wang2024emu3} & 8B & 85.2 & 1244 & 58.5 & 68.2 & 60.3 & 31.6 & 37.2 \\
    & InstructBLIP~\cite{dai2023instructblipgeneralpurposevisionlanguagemodels} & 13B & 78.9 & 1212.8 & - & - & 49.5 & - & 25.6 \\
        \hline
    \multirow{15}{*}{Und. and Gen.}& DreamLLM\textdagger~\cite{dong2023dreamllm} & 7B & - & - & - & - & - & - & 36.6 \\
    & LaViT\textdagger~\cite{jin2023unified} & 7B & - & - & - & - & 46.8 & - & - \\
    & MetaMorph\textdagger~\cite{tong2024metamorph} & 8B & - & - & 75.2 & 71.8 & - & - & - \\
    & NEXT-GPT\textdagger~\cite{wu2024next} & 13B & - & - & - & - & - & - & - \\
    \hdashline 
    & Show-o-256~\cite{xie2024show} & 1.3B & 73.8 & 948.4 & - & - & 48.7 & 25.1 & - \\
    & SDHo-512~\cite{xie2024show} & 1.3B & 80.0 & 1097.2 & - & - & 58.0 & 26.7 & - \\
    & D-DiT~\cite{li2024dual} & 2.0B & 84.0 & 1124.7 & - & - & 59.2 & - & - \\
    & Gemini-Nano-1~\cite{team2023gemini} & 1.8B & - & - & - & - & - & 26.3 & - \\
    & TokenFlow-XL~\cite{qu2024tokenflow} & 13B & 86.8 & \textbf{1545.9} & 68.9 & 68.7 & \textbf{62.7} & 38.7 & 40.7 \\
    & LWM~\cite{liu2024world} & 7B & 75.2 & - & - & 44.8 & - & - & 9.6 \\
    & VILA-U~\cite{wu2024vila} & 7B & 85.8 & 1401.8 & - & 59.0 & 60.8 & - & 33.5 \\
    & Chameleon~\cite{team2024chameleon} & 7B & - & - & - & - & - & 22.4 & 8.3 \\
    & Janus~\cite{wu2024janus} & 1.5B & 87.0 & 1338.0 & 69.4 &63.7 & 59.1 & 30.5 & 34.3 \\
    & JanusPro~\cite{chen2025janus} & 1.5B & 86.2 & 1444.0 & 75.5 & 68.3 & 59.3 & 36.3 & 39.8 \\
    & \textbf{UniDecouple (Ours)} & 3B& \textbf{87.2}& 1534.1& \textbf{79.3}& \textbf{72.4}& 61.5&\textbf{41.7}& \textbf{46.7}\\
    \hline
  \end{tabular}
  }
  \label{compare-understanding}
\end{table*}

\begin{table*}[htbp]
  \centering
  \setlength{\tabcolsep}{4pt} 
  \small 
  \caption{Evaluation of text-to-image generation ability on GenEval benchmark. "Und." and "Gen." denote "understanding" and "generation", respectively. Model using external pretrained diffusion model are marked with †.}
    \resizebox{1\linewidth}{!}{%
  \begin{tabular}{clccccccc}
    \hline
    \textbf{Type} & \textbf{Method} & \textbf{Single Obj.} & \textbf{Two Obj.} & \textbf{Counting} & \textbf{Colors} & \textbf{Position} & \textbf{Color Attri.} & \textbf{Overall} \\
    \hline
  \multirow{10}{*}{Gen. Only}      & LlamaGen~\cite{sun2024autoregressive} &0.71 & 0.34 & 0.21 & 0.58 & 0.07 & 0.04 & 0.32 \\
& LDM~\cite{rombach2022high} & 0.92 & 0.29 & 0.23 & 0.70 & 0.02 & 0.05 & 0.37\\
& SDv1.5~\cite{rombach2022high} & 0.97 & 0.38  & 0.35  & 0.76  & 0.04& 0.06 & 0.43  \\
& PixArt-$\alpha$~\cite{chen2023pixart} & 0.98 & 0.50 & 0.44 & 0.80 & 0.08 & 0.07 & 0.48 \\
 & SDv2.1~\cite{rombach2022high} & 0.98& 0.51 & 0.44 & 0.85& 0.07& 0.17 & 0.50 \\
& DALL-E2~\cite{ramesh2022hierarchical}    & 0.94 & 0.66& 0.49 & 0.77 & 0.10& 0.19 & 0.52 \\
& Emu3-Gen~\cite{wang2024emu3}    & 0.98  & 0.71   & 0.34  & 0.81 & 0.17 & 0.21 & 0.54 \\
& SDXL~\cite{podell2023sdxl}        & 0.98  & 0.74  & 0.39 & 0.85 & 0.15 & 0.23 & 0.55 \\
& DALL-E 3~\cite{betker2023improving}     & 0.96 & 0.87  & 0.47 & 0.83 & 0.43  & 0.45  & 0.67 \\
& SD3-Medium~\cite{esser2024scaling} & 0.99  & \textbf{0.94}  & \textbf{0.72}  & 0.89  & 0.33& 0.60& 0.74 \\
    \hline
   \multirow{12}{*}{Und. and Gen.}  & SEED-X†~\cite{li2024seed} & 0.97 & 0.58  & 0.26 & 0.80 & 0.19 & 0.14  & 0.49  \\
   \hdashline
& Show-o~\cite{xie2024show}    & 0.95  & 0.52 & 0.49& 0.82 & 0.11 & 0.28 & 0.53 \\
& D-DiT~\cite{li2024dual}     & 0.97& 0.80& 0.54& 0.76 & 0.32& 0.50& 0.65\\
& LWM~\cite{liu2024world}       & 0.93& 0.41& 0.46& 0.79& 0.09& 0.15& 0.47\\
& Transfusion~\cite{zhou2024transfusion} & - & - & -& - & - & - & 0.63\\
& ILLUME~\cite{wang2024illume}    & 0.99 & 0.86 & 0.45 & 0.71 & 0.39 & 0.28& 0.61  \\
& TokenFlow-XL~\cite{qu2024tokenflow} & 0.95& 0.60& 0.41& 0.81 & 0.16& 0.24& 0.53\\
& Chameleon~\cite{team2024chameleon}   & - & - & - & - & - & - & 0.39\\
& Janus~\cite{wu2024janus}       & 0.97& 0.68& 0.30& 0.84& 0.46 & 0.42 & 0.61 \\
& JanusPro~\cite{chen2025janus}     & 0.98 & 0.82 & 0.51& 0.89& \textbf{0.65}& 0.56 & 0.73\\
 & \textbf{UniDecouple (Ours)}& \textbf{0.99}& 0.84& 0.55& \textbf{0.91}& 0.62& \textbf{0.65}& \textbf{0.76}\\                    
    \hline
\end{tabular}}
\label{compare-generation1}
\end{table*}

\begin{table}[htbp]
  \centering
  \setlength{\tabcolsep}{6pt} 
  \small 
  \caption{Performances on DPG-Bench. "Und." and "Gen." denote "understanding" and "generation", respectively. Model using external pretrained diffusion model are marked with †.}
    \begin{tabular}{@{}lcccccc@{}}
    \hline
    Method & Global & Entity & Attribute & Relation & Other & Overall \\ 
    \hline
    SDv1.5 \cite{rombach2022high} & 74.63 & 74.23 & 75.39 & 73.49 & 67.81 & 63.18 \\
    PixArt-$\alpha$ \cite{chen2023pixart} & 74.97 & 79.32 & 78.60 & 82.57 & 76.96 & 71.11 \\
    SDXL \cite{podell2023sdxl} & 83.27 & 82.43 & 80.91 & 86.76 & 80.41 & 74.65 \\
    Hunyuan-DiT \cite{li2024hunyuan} & 84.59 & 80.59 & 88.01 & 74.36 & 86.41 & 78.87 \\
    PixArt-$\Sigma$ \cite{chen2024pixart} & 86.89 & 82.89 & 88.94 & 86.59 & 87.68 & 80.54 \\
    Emu3-Gen \cite{wang2024emu3} & 85.21 & 86.68 & 86.84 & 90.22 & 83.15 & 80.60 \\
    DALL-E 3 \cite{betker2023improving} & \textbf{90.97} & 89.61 & 88.39 & \textbf{90.58} & \textbf{89.83} & 83.50 \\
    SD3-Medium \cite{esser2024scaling} & 87.90 & \textbf{91.01} & 88.83 & 80.70 & 88.68 & 84.08 \\
    Janus~\cite{wu2024janus} & 82.33 & 87.38 & 87.70 & 85.46 & 86.41 & 79.68 \\
    JanusPro~\cite{chen2025janus} & 87.58 & 88.63 & 88.17 & 88.98 & 88.30 & 82.63 \\
    UniDecouple (Ours) & 87.95& 89.07& \textbf{89.31}& 89.33& 88.84&\textbf{84.11} \\
    \hline
\end{tabular}
\label{dpg}
\end{table}

All in all, our UniDecouple, based on~\cite{chen2025janus}, preserves strong MMU ability while achieving T2I quality comparable to state-of-the-art methods. This success is primarily attributed to TA-MoE, which effectively mitigates TOC between MMU and T2I, thereby significantly enhancing model’s overall capabilities.

\subsection{Qualitative Evaluation}

\begin{figure*}
    \centering
    \includegraphics[width=1\linewidth]{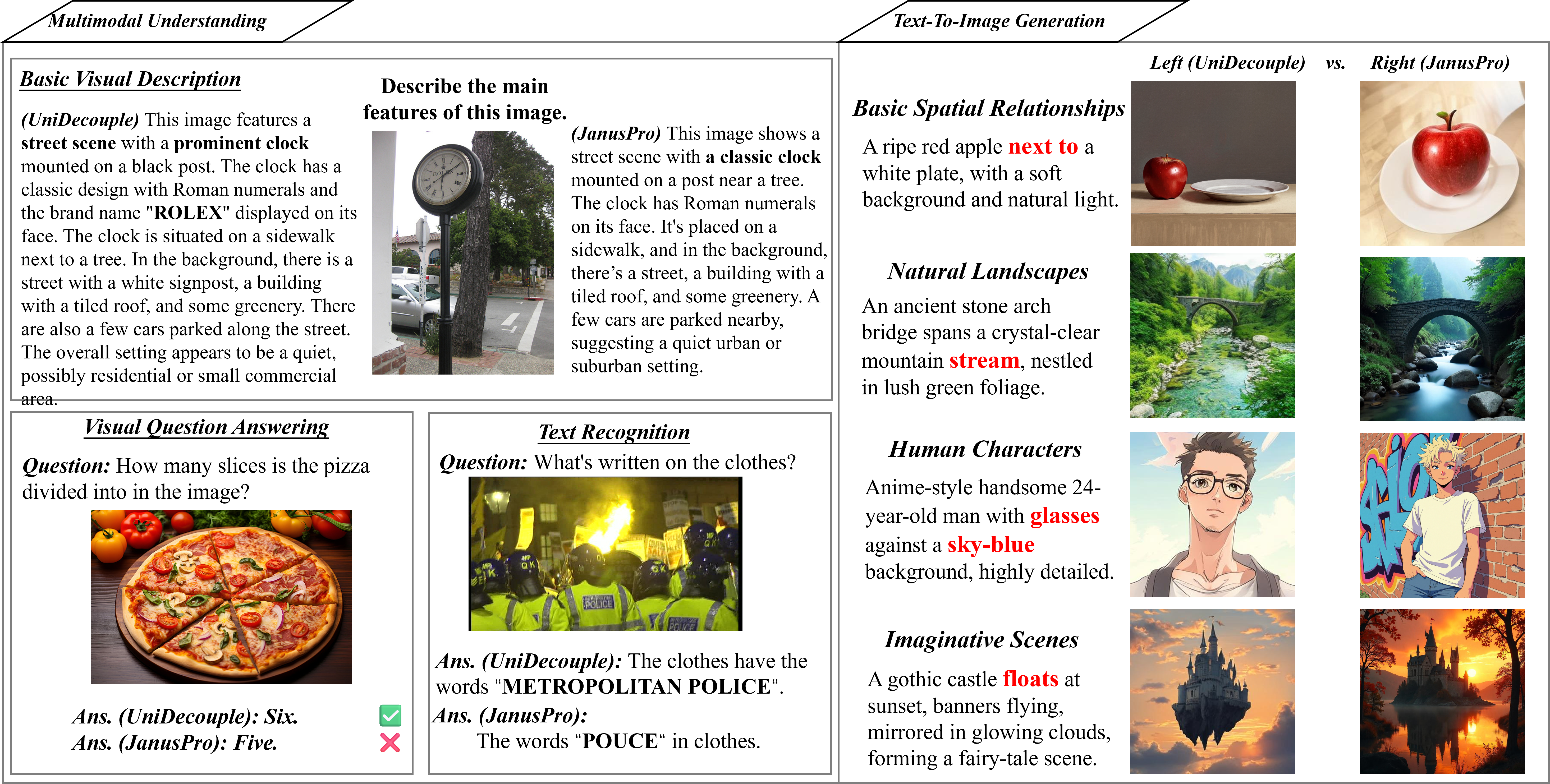}
    \caption{
    Qualitative comparison between UniDecouple and JanusPro on multimodal understanding and generation.}
    \label{fig:example}
\end{figure*}

\begin{figure}
    \centering
    \includegraphics[width=0.75\linewidth]{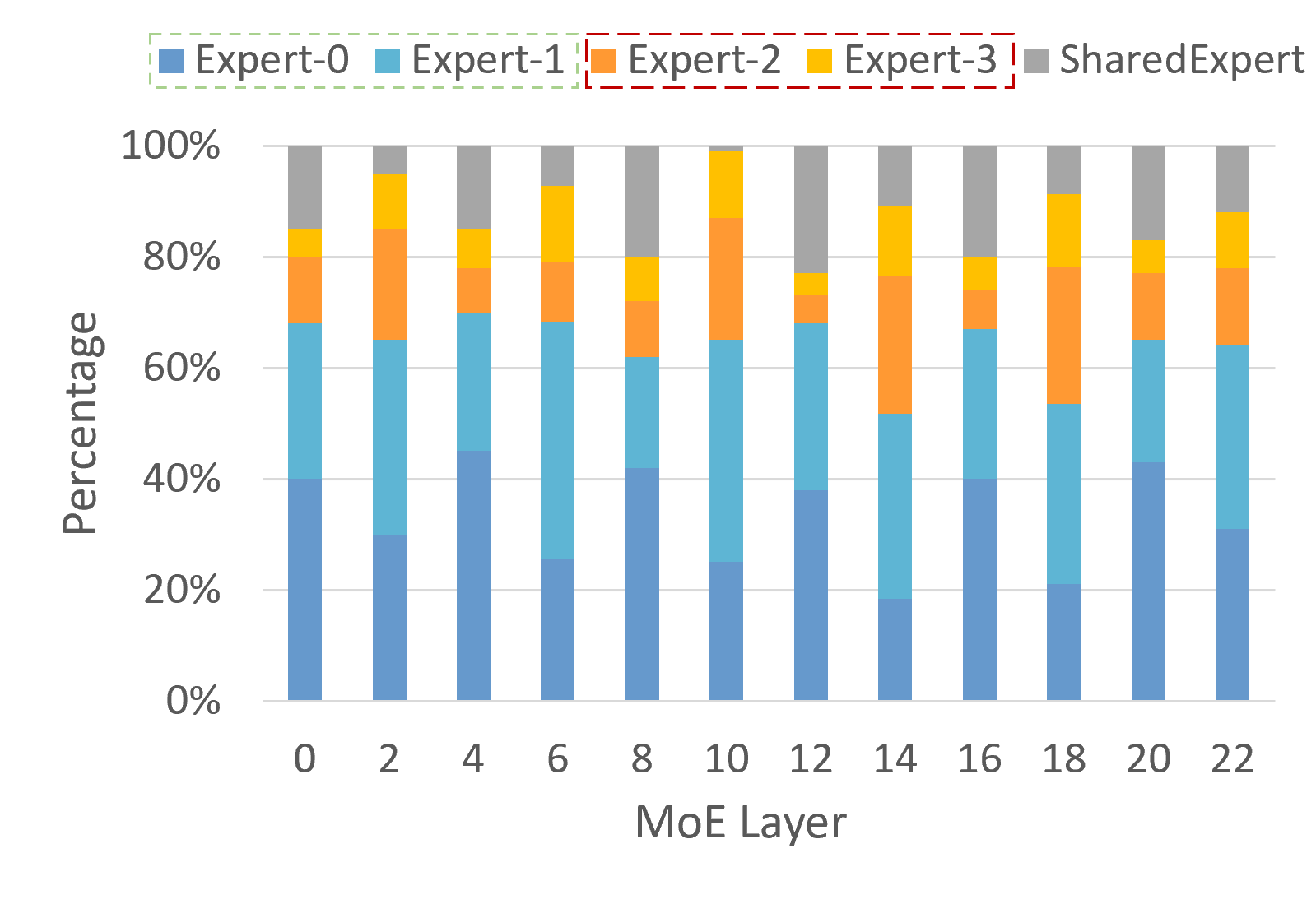}
    \caption{Visualization of the average expert load distribution based on a random sample of 100 instances from MMU.}
    \label{fig:expert-load}
\end{figure}

Figure~\ref{fig:example} presents qualitative results for both multimodal understanding (left) and text-to-image generation (right). UniDecouple demonstrates outstanding comprehension capabilities when processing inputs from diverse contexts, showcasing its strong capacity for multimodal semantic modeling, particularly in Basic Visual Description, Visual Question Answering (VQA), and Text Recognition tasks. The right side presents representative T2I results across four categories: basic spatial relationships, natural landscapes, human characters, and imaginative scenes. UniDecouple effectively captures the semantic intent of prompts, generating coherent, well-structured, and detail-rich images. These results highlight the superiority of our approach. UniDecouple not only models high-level semantic information in MMU tasks but also faithfully preserves fine-grained visual details in generation tasks, thereby mitigating inter-TOC.

\subsection{Expert Load Analysis}
\label{expert}

Fig.~\ref{fig:expert-load} illustrates expert utilization in understanding task, visualized by averaging values from randomly selected samples across multiple experimental runs. As shown in Fig.~\ref{fig:expert-load}, Expert-0 and Expert-1 belong to the understanding group, with their combined activation rate exceeding 60\% in most cases. Experts 2 and 3, belonging to the generation group, exhibit low activation rates across most layers, indicating that our task-aware router effectively routes tokens to the intended expert groups. Even when occasional routing errors increase generation group activations, the shared expert’s influence can partially compensate for meeting the task requirements. This further validates the effectiveness of our routing mechanism and shared expert design. 

\section{Ablation Study}
\label{ablation}

In this section, we conduct a comprehensive ablation study to investigate the contributions of key components in our proposed framework. We focus on analyzing the effectiveness of Task-Aware Mixture of Experts (TA-MoE), training strategy and expert ratio.

\begin{revision}
\subsection{Module-Level Separation for Continual Adaptation}
\label{sec:module_separation}

We examine where parameter specialization is most effective when continually adapting a model that already supports MMU and T2I. Starting from the same Janus-Pro checkpoint, we compare a shared baseline with attention-only, FFN-only, and combined attention--FFN separation. These variants use the same adaptation data, objectives, update counts, and evaluation protocols. For the three separated controls, branch selection is fixed by the ground-truth task identity, allowing us to compare separation locations without introducing learned routing. We additionally report the full UniDecouple model with its shared expert and learnable token-level router. These controls examine architectural choices inspired by decoupled unified models, rather than reproduce BAGEL, MoT, LMFusion, or Mogao.

\begin{table*}[t]
\centering
\begin{revision}
\small
\setlength{\tabcolsep}{4pt}
\captionsetup{labelfont={color=black},textfont={color=black}}
\caption{Ablation of module-level separation. All variants except Janus-Pro are initialized from the same Janus-Pro checkpoint and use identical training data, objectives, update counts, and evaluation protocols.}
\label{tab:module_separation}
\resizebox{\textwidth}{!}{%
\begin{tabular}{lcccccccc}
\toprule
Variant & Attn. split & FFN split & Routing &
MMMU$\uparrow$ & POPE$\uparrow$ & MM-Vet$\uparrow$ &
GenEval$\uparrow$ & DPG-Bench$\uparrow$ \\
\midrule
JanusPro
& No & No & -- &
36.3 & 86.2 & 39.8 &
0.73 & 82.63 \\

Attention-only separation
& Yes & No & Fixed task & 36.4 & 86.2 & 39.7 & 0.71 & 82.66 \\

FFN-only separation
& No & Yes & Fixed task & 36.9 & 86.4 & 41.8 & 0.73 & 82.68 \\

Attention + FFN separation
& Yes & Yes & Fixed task & 35.7 & 85.7 & 39.1 & 0.69 & 81.52 \\

UniDecouple
& No & Yes & Learned token &
\textbf{41.7} & \textbf{87.2} & \textbf{46.7} &
\textbf{0.76} & \textbf{84.11} \\
\bottomrule
\end{tabular}%
}
\end{revision}
\end{table*}

Table~\ref{tab:module_separation} shows that FFN-only separation performs best among the three fixed-routing separation strategies on all five benchmarks. Relative to the shared baseline with matched adaptation, it improves MMMU from $36.3$ to $36.9$, POPE from $86.2$ to $86.4$, MM-Vet from $39.8$ to $41.8$, and DPG-Bench from $82.63$ to $82.68$, while maintaining a GenEval score of $0.73$. Adding attention separation to the FFN-only design reduces MMMU from $36.9$ to $35.7$, POPE from $86.4$ to $85.7$, MM-Vet from $41.8$ to $39.1$, GenEval from $0.73$ to $0.69$, and DPG-Bench from $82.68$ to $81.52$. Thus, specializing additional parameter blocks does not necessarily improve performance within this adaptation schedule.

The full UniDecouple model further improves over FFN-only fixed routing on all five benchmarks, reaching $41.7$ on MMMU, $87.2$ on POPE, $46.7$ on MM-Vet, $0.76$ on GenEval, and $84.11$ on DPG-Bench. This comparison supports the combined design of specialized FFNs, an explicit shared expert, and learned token-level routing. The component and routing ablations in Tables~\ref{tab:albation1} and~\ref{tab:routing_comparison} further examine their respective contributions.

Overall, these results favor retaining shared attention and selectively specializing FFNs over jointly separating attention and FFNs in our continual adaptation setting. One plausible explanation is that separate attention branches require more diverse data or a longer optimization schedule to realize the benefits of task-specific dependency modeling. With limited additional data and updates, shared attention may better reuse the pretrained model's contextual computation while FFN specialization accommodates task-dependent transformations.
\end{revision}

\subsection{Effectiveness of TA-MoE}

\begin{table}[ht]
  \centering
  \small
  \caption{Ablation study on Task-Aware Router (incl. Group Loss) and Shared Expert.}
  \label{tab:albation1}
  \begin{tabular}{cccccclc}
    \hline
    \multirow{2}{*}{\textbf{Model}} & \multicolumn{2}{c}{\textbf{Task-Aware Router}} & \textbf{Shared}  & \multicolumn{3}{c}{\textbf{Understanding}}&\textbf{Generation}\\
    & \multicolumn{2}{c}{\textbf{+ Group Loss}} & \textbf{Expert}  & MMMU & & POPE&Gen.eval\\
    \hline
    A & \multicolumn{2}{c}{\ding{55}} & \ding{55}  & 36.8 & & 85.4&0.70 \\
    B & \multicolumn{2}{c}{\checkmark} & \ding{55}  & 37.6 & & 85.9&0.73 \\
    C & \multicolumn{2}{c}{\checkmark} & \checkmark  & \textbf{41.7} & & \textbf{87.2}  &\textbf{0.76} \\
    \hline
  \end{tabular}
\end{table}

To assess the effectiveness of each component in the TA-MoE, we progressively construct a series of ablation models, as shown in Table~\ref{tab:albation1}. We begin with Model A as the baseline, which excludes two critical components: the Task-Aware Router (including Group Loss) and the Shared Expert. Model B introduces the Task-Aware Router, enabling dynamic expert group selection based on task-specific requirements. Building on this, Model C incorporates the Shared Expert module, which promotes cross-task knowledge sharing and enhances the model’s generalization capability.

We evaluate understanding performance using the MMMU and POPE benchmarks, and assess generation quality with the Gen.eval metric, reporting the “overall” score as the primary evaluation criterion. As shown in Table~\ref{tab:albation1}, each added component yields consistent performance gains, with Model C achieving the best results across all tasks. This ablation demonstrates the effectiveness and necessity of each module in enhancing both task-specific adaptation and generalization capabilities within the TA-MoE framework.

\subsection{Effectiveness of Training Strategy}

We investigate the impact of different training strategies on model performance through two distinct model variants presented in Table~\ref{tab:albation2}. To ensure a fair comparison, all models share the same architecture and are evaluated on identical task sets. The comparative results between models D and E demonstrate that the two-stage training strategy employed by model E yields significant improvements across various benchmarks.  This indicates that incorporating dedicated training stages for handling distinct tasks not only enhances model performance but also accelerates convergence, underscoring the strategic advantage of involving experts in task-specific training.

\begin{table}[ht]
  \centering
  \footnotesize
  \caption{\revisiontext{Ablation study of the proposed two-stage training strategy and Stage-1 expert initialization.}}
  \label{tab:albation2}
  \begin{tabular}{l c c c c c c c}
    \toprule
    \textbf{Model} & \textbf{MoE} & \textbf{Training} & \textbf{Expert Init.} 
    & \textbf{Epochs} $\downarrow$ & \textbf{MMMU} $\uparrow$ 
    & \textbf{POPE} $\uparrow$ & \textbf{GenEval} $\uparrow$ \\
    \midrule
    D & \checkmark & Single-Stage & Pure & 94 & 36.5 & 85.3 & 0.68 \\
    E & \checkmark & Two-Stage & Stage 1 & 76 & \textbf{41.7} & \textbf{87.2} & \textbf{0.76} \\
    \bottomrule
   \end{tabular}
\end{table}

\subsection{Impact of Expert Ratio}

\begin{revision}
To examine the balance between task specialization and cross-task knowledge sharing, Table~\ref{tab:ratio} compares different group-specific-to-shared expert (G:S) ratios on POPE, GenEval, MMMU, MM-Vet, and DPG-Bench, with POPE further decomposed into Random, Popular, and Adversarial subsets. All results are averaged over three independent runs under the same experimental configuration apart from the expert ratio. The $1{:}0$ and $0{:}1$ configurations use only group-specific experts and only a shared expert, respectively, while the remaining configurations combine both types.
\end{revision}

\begin{table}[ht]
  \centering
  \begin{revision}
  \footnotesize
  \setlength{\tabcolsep}{5pt}
  \captionsetup{labelfont={color=black},textfont={color=black}}
  \caption{Extended evaluation of different group-specific-to-shared expert (G:S) ratios. Results are averaged over three independent runs. Bold indicates the best result in each column.}
  \label{tab:ratio}
  \begin{tabular}{@{}lcccccccc@{}}
    \toprule
    \multirow{2}{*}{G:S}
    & \multicolumn{4}{c}{POPE}
    & \multirow{2}{*}{GenEval}
    & \multirow{2}{*}{MMMU}
    & \multirow{2}{*}{MM-Vet}
    & \multirow{2}{*}{DPG-Bench} \\
    \cmidrule(lr){2-5}
    & Overall & Random & Popular & Adversarial
    & & & & \\
    \midrule
    $1{:}0$ & 85.9 & 86.8 & 85.1 & 85.8 & 0.73 & 37.6 & 42.1 & 82.77 \\
    $0{:}1$ & 86.7 & 86.9 & \textbf{88.0} & 85.2 & 0.74 & 37.4 & 41.4 & 82.35 \\
    $1{:}1$ & 87.1 & 87.5 & 87.8 & 86.0 & 0.74 & 38.5 & 43.6 & 83.44 \\
    $2{:}1$ & \textbf{87.2} & \textbf{88.0} & 87.7 & 85.9 & \textbf{0.76} & \textbf{41.7} & \textbf{46.7} & \textbf{84.11} \\
    $3{:}1$ & 86.9 & 87.6 & 87.5 & 85.6 & 0.73 & 39.2 & 44.0 & 82.89 \\
    \bottomrule
  \end{tabular}
  \end{revision}
\end{table}

\begin{revision}
The shared-only configuration retains most of the performance of the $2{:}1$ hybrid on overall POPE and GenEval ($86.7$ vs.\ $87.2$ and $0.74$ vs.\ $0.76$), indicating that shared computation remains valuable. Conversely, removing the shared expert reduces performance on these two benchmarks, although the group-specific-only configuration outperforms the shared-only configuration on MMMU ($37.6$ vs.\ $37.4$) and MM-Vet ($42.1$ vs.\ $41.4$). Thus, separating the expert pathways alone does not consistently improve downstream performance. As discussed in Sec.~\ref{subsec:conflict}, cross-task gradient interactions can be either conflicting or complementary; TOC does not imply that MMU and T2I conflict across all parameters and computations. The observed results are consistent with a trade-off: task-specific pathways can reduce interference, while removing the shared expert can limit feature reuse and knowledge transfer through expert layers.

The $2{:}1$ hybrid achieves the highest overall scores on all five benchmarks, with larger gains over the shared-only configuration on MMMU ($+4.3$), MM-Vet ($+5.3$), and DPG-Bench ($+1.76$). Its POPE gains vary by subset: Random improves by $1.1$ points, Adversarial by $0.7$ points, and Popular decreases by $0.3$ points. These differences show that the modest overall POPE improvement does not fully describe the benefits across evaluation settings. The results support combining task-specialized and shared computation to balance specialization and knowledge sharing, rather than treating complete decoupling as sufficient to resolve TOC. Downstream scores alone do not establish the presence or elimination of gradient conflict in any configuration.

Increasing the G:S ratio from $2{:}1$ to $3{:}1$ reduces performance on all five benchmarks and all three POPE subsets. This is consistent with diminishing benefits from additional specialization relative to shared computation and supports $2{:}1$ as the best ratio among those tested under our training setup. Due to differences in routing strategy, we do not directly adopt the $3{:}1$ ratio reported in DeepSeek~\cite{dai2024deepseekmoe}; the preferred ratio may depend on the architecture and training configuration.
\end{revision}

\begin{revision}
\subsection{Comparison of Routing Strategies}
\label{sec:routing_ablation}

\begin{table}[htbp]
\centering
\begin{revision}
\footnotesize
\setlength{\tabcolsep}{4pt}
\captionsetup{labelfont={color=black},textfont={color=black}}
\caption{Comparison of different routing strategies. Oracle hard routing uses the ground-truth sample-level task identity, whereas the proposed router makes token-level decisions from the current hidden representations.}
\label{tab:routing_comparison}
\begin{tabular}{lccccc}
\toprule
Routing strategy & Grouping accuracy & POPE & GenEval & MMMU & DPG-Bench \\
\midrule
Standard Top-2 MoE
& -- & $84.6$ & $0.63$ & $35.7$ & $81.72$ \\
Hard routing w/ Shared Expert
& $100\%$ & $86.8$ & $0.74$ & $37.2$ & $83.03$ \\
Hard routing w/o Shared Expert
& $100\%$ & $86.4$ & $0.73$ & $36.9$ & $82.68$ \\
Ours: learnable token-level routing
& $\sim76\%$ & $\mathbf{87.2}$ & $\mathbf{0.76}$ & $41.7$ & $84.11$ \\
\bottomrule
\end{tabular}
\end{revision}
\end{table}

The sample-level task identity specifies the global objective of an input, but does not uniquely determine the computation required by every token at every layer. We therefore compare learnable token-level routing with oracle task-ID routing and standard Top-2 MoE. Oracle routing assigns all tokens in a sample to the expert group indicated by its ground-truth task identity. We evaluate this baseline both with and without the shared expert. Standard Top-2 MoE selects experts without the explicit task-group structure.

We define grouping accuracy as the fraction of token-level group selections that agree with the sample-level task identity, excluding the always-active shared expert. Oracle routing achieves $100\%$ by construction, whereas the learned router achieves approximately $76\%$, corresponding to approximately $24\%$ cross-task group selections. As specified in Sec.~\ref{tamoe}, the learned router is trained with an auxiliary task-supervised group loss. This loss encourages agreement with the task label, but does not impose the deterministic assignment used by oracle routing: the actual group selection is predicted from each token's hidden representation.

Table~\ref{tab:routing_comparison} shows that learned routing achieves a POPE score of $87.2$ and a GenEval score of $0.76$, compared with $86.8$ and $0.74$ for oracle routing with the shared expert, and $84.6$ and $0.63$ for standard Top-2 MoE. The oracle baseline with the shared expert is the closer routing control because it retains both a task-specific expert and the shared expert. Removing the shared expert additionally changes the active expert configuration, yielding a POPE score of $86.4$ and a GenEval score of $0.73$.

Thus, perfect agreement with the sample-level task label does not guarantee better downstream performance. The results are consistent with useful cross-task computation and motivate retaining token-dependent group selection. They do not establish that every cross-task selection is beneficial. In particular, the MMU expert-load visualization in Fig.~\ref{fig:expert-load} describes routing behavior; the oracle comparison provides the corresponding downstream evaluation.
\end{revision}

\begin{revision}
\subsection{Model Capacity and Activated Parameters}
\label{sec:capacity_ablation}

\begin{table}[htbp]
\centering
\begin{revision}
\small
\setlength{\tabcolsep}{4pt}
\captionsetup{labelfont={color=black},textfont={color=black}}
\caption{Model capacity and benchmark performance. Parameter counts are in billions. Both UniDecouple variants activate one task-specific expert and one shared expert per token.}
\label{tab:parameter_comparison}
\begin{tabular}{lccccccc}
\toprule
Model & $d_{\mathrm{ff}}$ & Total & Activated & Active Experts
& MMMU & GenEval & POPE \\
\midrule
Janus-Pro
& 5632 & 1.500B & 1.500B & Dense
& 36.3 & 0.73 & 86.2 \\
UniDecouple-3B
& 2880 & 2.795B & 1.520B & $1+1$
& 41.7 & 0.76 & 87.2 \\
UniDecouple-5B
& 5632 & 4.820B & 2.330B & $1+1$
& 43.0 & 0.77 & 87.9 \\
\bottomrule
\end{tabular}
\end{revision}
\end{table}

The approximately $3\mathrm{B}$ model size of UniDecouple denotes total stored parameters, rather than the parameters activated for each token. As reported in Table~\ref{tab:parameter_comparison}, UniDecouple-3B stores approximately $2.795\mathrm{B}$ parameters and activates $1.520\mathrm{B}$ per token, including one task-specific expert and one shared expert. Dense Janus-Pro stores and activates approximately $1.500\mathrm{B}$ parameters. Each UniDecouple-3B expert has an FFN intermediate dimension of $2880$, compared with $5632$ for Janus-Pro.

Under these comparable activated parameter budgets, UniDecouple-3B improves MMMU from $36.3$ to $41.7$, POPE from $86.2$ to $87.2$, and GenEval from $0.73$ to $0.76$. This comparison distinguishes stored expert capacity from per-token activated capacity. It does not match total stored capacity, and activated parameter counts alone do not establish identical FLOPs or inference latency.

To examine FFN width, we additionally train UniDecouple-5B with $d_{\mathrm{ff}}=5632$, matching the FFN width of Janus-Pro while keeping the remaining UniDecouple training settings unchanged. This variant stores $4.820\mathrm{B}$ parameters and activates $2.330\mathrm{B}$ per token, achieving $43.0$ on MMMU, $87.9$ on POPE, and $0.77$ on GenEval. Its improvement over UniDecouple-3B accompanies increases in both total and activated parameter counts and therefore cannot be attributed solely to routing.
\end{revision}

\begin{revision}
\subsection{Data-Matched Fine-Tuning Control}
\label{sec:data_matched_ablation}

\begin{table}[htbp]
\centering
\begin{revision}
\small
\setlength{\tabcolsep}{3.5pt}
\captionsetup{labelfont={color=black},textfont={color=black}}
\caption{Data-matched comparison with Janus-Pro. ``Same data/LoRA'' denotes adaptation on ShareGPT4V $+$ MidJourney using the same data sampling, training schedule, optimizer, and rank-16 LoRA configuration as UniDecouple.}
\label{tab:r3_matched_control}
\begin{tabular}{@{}lccccc@{}}
\toprule
Model & Matched adaptation & MoE & MMMU & POPE & GenEval \\
\midrule
Janus-Pro (original)
& No & No & $36.3$ & $86.2$ & $0.73$ \\
Janus-Pro + same data/LoRA
& Yes & No & $37.4$ & $86.7$ & $0.74$ \\
UniDecouple
& Yes & Yes & $\mathbf{41.7}$ & $\mathbf{87.2}$ & $\mathbf{0.76}$ \\
\bottomrule
\end{tabular}
\end{revision}
\end{table}

To assess whether the improvement is explained by additional fine-tuning data, we adapt Janus-Pro on the same ShareGPT4V and MidJourney data used for UniDecouple, with the same sampling ratio, number of training steps, optimizer, and rank-16 LoRA configuration. We retain the original dense FFNs and introduce no MoE or expert-routing module.

As shown in Table~\ref{tab:r3_matched_control}, this adaptation improves Janus-Pro from $36.3$ to $37.4$ on MMMU, from $86.2$ to $86.7$ on POPE, and from $0.73$ to $0.74$ on GenEval. Under the same data and LoRA adaptation protocol, UniDecouple achieves $41.7$, $87.2$, and $0.76$, respectively. The remaining gains of $4.3$ points on MMMU, $0.5$ points on POPE, and $0.02$ on GenEval indicate that the additional data and LoRA adaptation alone do not explain the observed improvement.

Together with the routing and parameter comparisons, these results support the benefit of combining shared and task-specialized experts with learned token-level routing under a comparable activated parameter budget. They do not fully isolate expert organization from the larger total stored capacity of UniDecouple.
\end{revision}
 
\section{Conclusion}

In this paper, we address \revisiontext{Task Objective Conflict (TOC) between multimodal understanding (MMU) and text-to-image generation (T2I) in unified autoregressive models under continual adaptation. Our theoretical and empirical analyses reveal task-level optimization interference alongside complementary token-level interactions, motivating selective task specialization with continued cross-task knowledge sharing.} We propose UniDecouple, \revisiontext{a selective decoupling framework that retains shared attention while introducing task-specific FFN experts to mitigate interference.} Central to our approach is the TA-MoE, which combines Hierarchical Expert Routing and Hybrid Expert Collaboration, \revisiontext{enabling adaptive token-level expert selection and knowledge transfer through shared experts. A two-stage training strategy further balances task specialization and coordinated adaptation.} Extensive experiments \revisiontext{and ablation studies demonstrate improvements in both multimodal understanding and image generation under a comparable per-token activated parameter budget, although total stored capacity increases. These results support selectively specializing FFNs while preserving shared computation as a promising approach to balancing knowledge preservation and task adaptation in unified multimodal models.}

\section*{Declarations}

\noindent \textbf{Funding.}
This work is supported by the Fundamental Research Funds for the Central Universities, Peking University.

\noindent \textbf{Conflict of Interest.}
The authors have no relevant financial or non-financial interests to disclose.

\noindent \textbf{Data Availability.}
The datasets analyzed during the current study are available in public repositories. Specifically:
The ShareGPT4V dataset for multimodal understanding.
The Midjourney Prompts dataset  for text-to-image generation, accessible at \url{https://huggingface.co/datasets/vivym/midjourney-prompts}.
All evaluation benchmarks used, including \revisiontext{POPE~\cite{li2023evaluating}, MMBench~\cite{liu2024mmbench}, and MMMU~\cite{yue2024mmmu}}, are publicly available research datasets. The code and models for UniDecouple will be made available on GitHub following the publication of this work.

\bibliography{sn-bibliography}

\end{document}